\documentclass{bmvc2k}

\usepackage{multirow}
\usepackage{booktabs}
\usepackage{caption}

\usepackage{breakurl}

\title{Blocks as Probes: Dissecting Categorization Ability of Large Multimodal Models}

\addauthor{Bin Fu}{bin.fu@vipl.ict.ac.cn}{1,2}
\addauthor{Qiyang Wan}{qiyang.wan@vipl.ict.ac.cn}{1,2}
\addauthor{Jialin Li}{jialin.li@vipl.ict.ac.cn}{1,2}
\addauthor{Ruiping Wang}{wangruiping@ict.ac.cn}{1,2}
\addauthor{Xilin Chen}{xlchen@ict.ac.cn}{1,2}

\addinstitution{
Key Laboratory of AI Safety of CAS,\\
Institute of Computing Technology,\\ Chinese Academy of Sciences (CAS)\\
Beijing, China
}
\addinstitution{
University of Chinese\\ Academy of Sciences \\
Beijing, China
}

\runninghead{Fu, Wan, Li, Wang, Chen}{Blocks as Probes}

\newcommand{\blue}[1]{\textcolor{blue}{#1}}

\begin{document}
\maketitle

\begin{abstract}

Categorization, a core cognitive ability in humans that organizes objects based on common features, is essential to cognitive science as well as computer vision. To evaluate the categorization ability of visual AI models, various proxy tasks on recognition from datasets to open world scenarios have been proposed. Recent development of Large Multimodal Models (LMMs) has demonstrated impressive results in high-level visual tasks, such as visual question answering, video temporal reasoning, etc., utilizing the advanced architectures and large-scale multimodal instruction tuning. Previous researchers have developed holistic benchmarks to measure the high-level visual capability of LMMs, but there is still a lack of pure and in-depth quantitative evaluation of the most fundamental categorization ability. According to the research on human cognitive process, categorization can be seen as including two parts: category learning and category use. Inspired by this, we propose a novel, challenging, and efficient benchmark based on composite blocks, called \textbf{ComBo}, which provides a disentangled evaluation framework and covers the entire categorization process from learning to use. By analyzing the results of multiple evaluation tasks, we find that although LMMs exhibit acceptable generalization ability in learning new categories, there are still gaps compared to humans in many ways, such as fine-grained perception of spatial relationship and abstract category understanding. Through the study of categorization, we can provide inspiration for the further development of LMMs in terms of interpretability and generalization.

\end{abstract}

\section{Introduction}
\label{sec:intro}

\begin{figure*}[t]
\begin{center}
\includegraphics[width=1\linewidth]{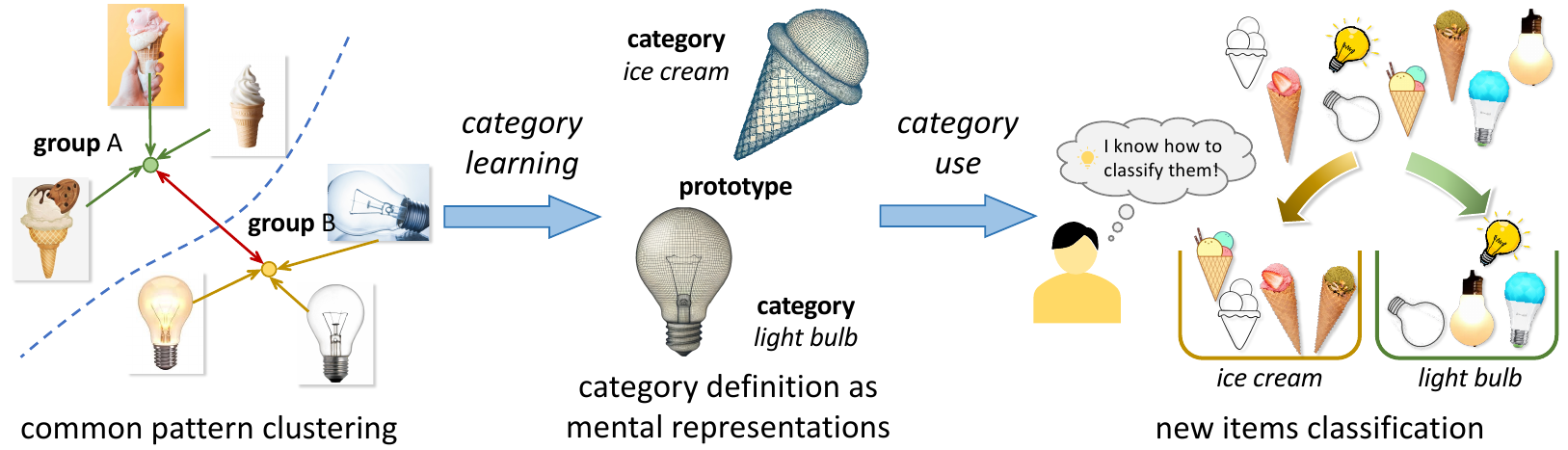}
\end{center}
\caption{Human behavior in  categorization. People can group objects together based on common patterns, form mental representation of categories, and classify novel items.}
\label{fig:human-categorization-abilities}
\vspace{-0.8em}
\end{figure*}

Categorization is one of the most fundamental cognitive abilities of humans. As shown in Fig.\ref{fig:human-categorization-abilities}, visual categorization involves the process of organizing objects into categories based on shared features or attributes (\textit{category learning}), and using the mental representation to complete cognitive tasks, such as classifying new objects (\textit{category use}) \cite{markman2003category}. The learning and use of categories is not only a significant research topic in cognitive science but is also considered a critical feature of artificial intelligence \cite{fei2022searching}.


With the progressive enhancement of computer vision models, there should be an implicit improvement in the capability for categorization, evaluated by the development of diverse proxy tasks, such as object recognition. In recent years, multimodal models like CLIP \cite{radford2021learning} align visual and textual modalities, thereby liberating categorization from the constraint of datasets and advancing towards open-world scenarios. Moreover, Large Multimodal Models (LMMs) have integrated modalities such as vision into language models with a large number of parameters, displaying remarkable performance on numerous high-level visual tasks \cite{singh2019towards, ye2023mplug} and holistic benchmarks \cite{yu2023mm, yue2023mmmu}. 
Excellent understanding shown in image captioning \cite{chen2015microsoft} and visual question answering \cite{goyal2017making, hudson2019gqa, marino2019ok} implies that LMMs seem to possess sufficient categorization ability \cite{wu2023gpt4vis}. 
However, there is a lack of direct, objective, and decoupled evaluations of LMMs' capabilities in the most fundamental tasks of visual perception.

To explore this question, a pure and in-depth benchmark is required to dissect categorization ability of LMMs.
We argue that an effective benchmark should have the following characteristics:
(1) \textbf{Avoiding data leakage}. Prevent not only data sample leakage \cite{fu2023challenger, mao2023gpteval} but also leakage of evaluation categories. Similar to using abstract reasoning to test human intelligence in Wechsler Adult Intelligence Scale \cite{wechsler1955wechsler}, some abstract and novel categories that are impossible to exist in the training set should be introduced.
(2) \textbf{Establishing quantitive and discriminative tasks}. Select diverse and quantifiable evaluation tasks and questions to ensure objectivity and maximize dissection efficiency, allowing us to explore the boundaries of their capability through failure case analysis \cite{fu2023challenger}.
(3) \textbf{Performing unit tests and integration tests}. Design a diverse set of evaluation tasks that cover the entire cognitive process of category learning and use. These tasks should avoid unrestrained end-to-end questions to ensure a disentangled evaluation.
To respond to these requirements, we will revisit the human cognitive process of categorization \cite{markman2003category, rosch1978principles, tversky1977features}, and attempt to decouple the key evaluation points to probe the categorization process of LMMs, which will be detailed in Sec.\ref{sec:dissection}.

To meet the aforementioned requirements of categorization evaluation, we construct a synthetic dataset consisting of \textbf{Com}posite \textbf{B}l\textbf{o}cks (\textbf{ComBo}), which will be elaborated in Sec.\ref{sec:combo}.
The objects and categories in ComBo are entirely unseen to LMMs, meeting the need of preventing data leakage.
Since the synthetic data is completely controllable, we can easily control the difficulty of the tasks and inexpensively generate a large number of questions with ground truth for quantitative evaluation.
Inspired by the analysis of the categorization process, we design a series of tasks covering the entire cognitive process, aiming to comprehensively evaluate the categorization capability of LMMs. First, we evaluate LMMs' ability to perceive low-level patterns that is critical for accurate object recognition. Next, we explore their capability to align abstract category representations with human mental concepts by predefined semantic categories, which verifies the consistency of the learned concepts. Finally, we challenge the models with unseen abstract categories to examine their generalizability of categorization ability. These experiments are designed to illuminate the strengths and limitations of LMMs in replicating human-like category cognition, thereby pushing the boundaries of LMMs in understanding and interacting with the real-world objects.

\begin{figure*}[t]
\begin{center}
\includegraphics[width=1\linewidth]{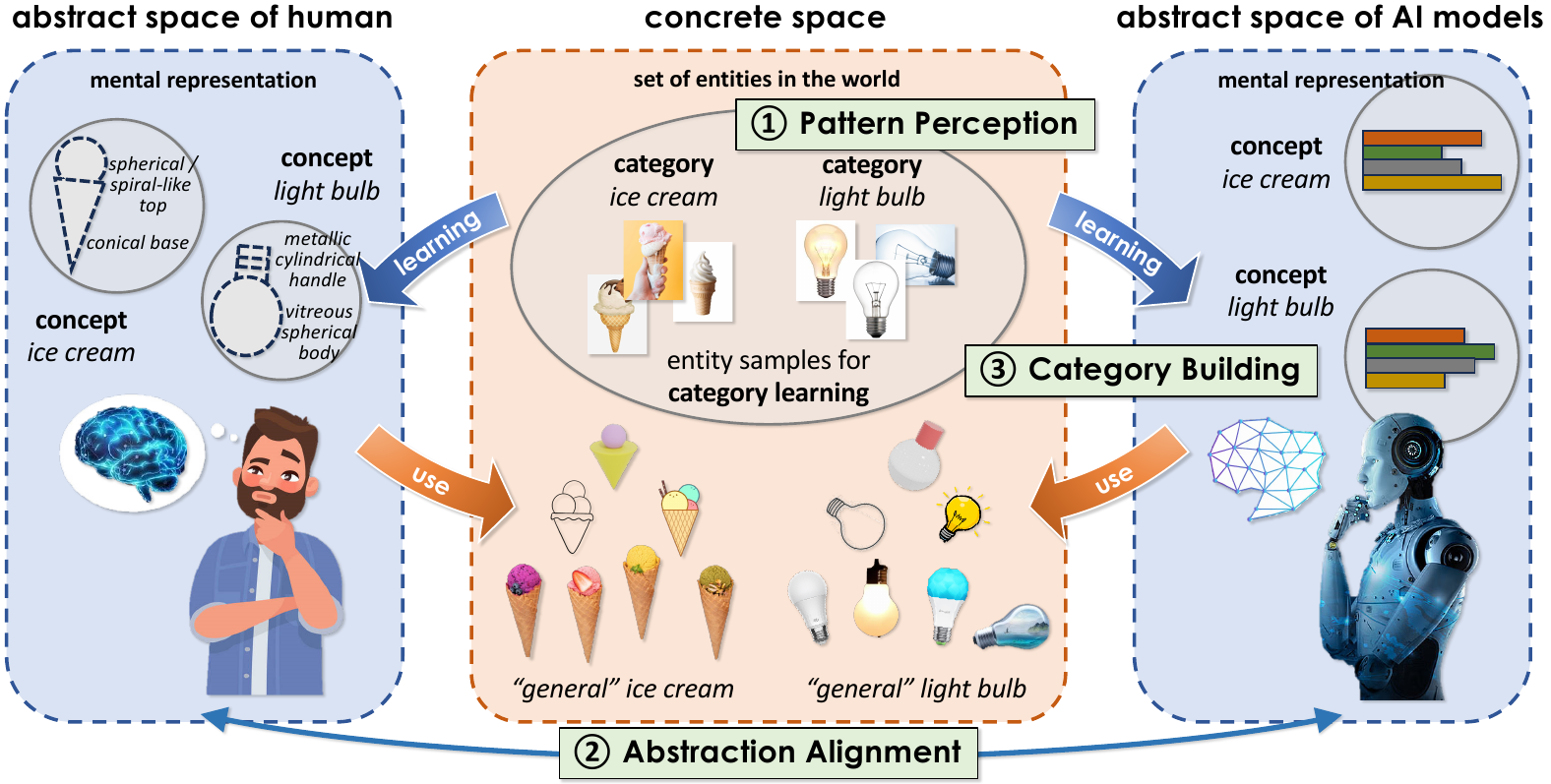}
\end{center}
\caption{The cognitive processes of humans and LMMs in categorization. Categorization can be modeled as a process of category learning and category use between concrete and abstract spaces. The proposed evaluation tasks are shown in green blocks.
}
\label{fig:cognitive-process}
\vspace{-0.8em}
\end{figure*}

The experimental results and analyses in Sec.\ref{sec:exp} reveal that while LMMs demonstrate enhanced categorization capability over traditional CV models, they continue to be stuck in spatial detail recognition, abstract conceptual reasoning, and learning unseen categories in some scenarios.
The corresponding discussions and related work are presented in Sec.\ref{sec:disussion} and Sec.\ref{sec:rw}.
Analyzing these failure cases allow us to explore LMMs' shortcomings from a more basic level and make effective promotions.
We believe that studying the low-level visual capability such as categorization of LMMs will contribute to the further development of generalizability and interpretability in AI models.

\section{Categorization Dissection}
\label{sec:dissection}

In this section, we will break down the design of categorization dissection, starting with an introduction to the cognitive processes of categorization of humans and LMMs in Sec.\ref{sec:overview-cat}, and based on this, we will introduce the design philosophy of our evaluation in Sec.\ref{sec:design-philosophy}.

\subsection{Overview of Categorization}
\label{sec:overview-cat}

As shown in Fig.\ref{fig:cognitive-process}, the cognitive process of categorization involves information transmission between \textit{concrete space} and \textit{abstract space} \cite{markman2003category}.
The concrete space consists of perceivable visual entities in the real world, including various data forms of the object categories, such as a photorealistic ice cream, an ice cream sketch, and toy blocks like an ice cream.
The abstract space is where both humans and LMMs store categorization rules about these categories respectively, such as the shape of the category ``ice cream'' (typically consisting of a cone and ice cream balls) and some attributes (a cold dessert).
Humans use mental representations to encode key aspects about category members \cite{markman2003category}, while LMMs store knowledge about entities in their internal implicit representation spaces, such as feature vectors.

Thus, the cognitive process of categorization can be represented as follows. (1) \textit{category learning}: humans and LMMs perceive data in concrete space, gathering some items with common features together. They then abstract and summarize the commonalities of these items to form a concept representation of the category in abstract space. (2) \textit{category use}: humans and LMMs utilize the concepts in abstract space to construct various cognitive functions. For example, classifying a newly encountered object as ice cream, inferring the cold taste and other attributes a new ice cream should have, and even implicitly applying it to tasks such as image captioning and visual question answering about a dessert shop.


\subsection{Design Philosophy of Evaluation}
\label{sec:design-philosophy}

Based on the cognitive process of object categorization described above, we design three evaluation tasks (green blocks in Fig.\ref{fig:cognitive-process}) corresponding to different stages of the categorization process, in order to conduct a comprehensive evaluation of the categorization capability of LMMs.

\begin{figure*}[t]
\begin{center}
\includegraphics[width=1\linewidth]{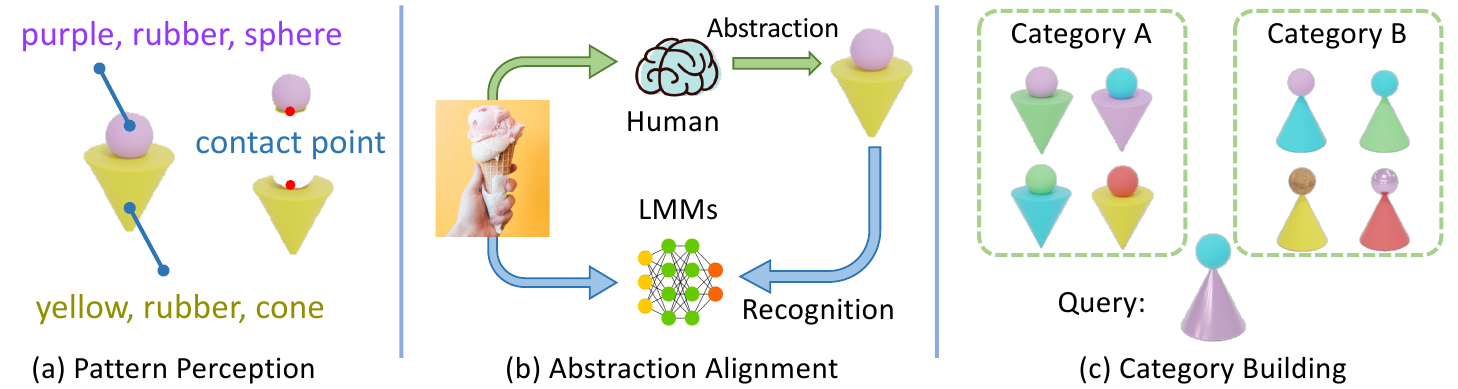}
\end{center}
\caption{Three progressive tasks on categorization evaluation. (a) \textbf{Pattern Perception}: Evaluating LMMs' low-level pattern recognition ability. (b) \textbf{Abstraction Alignment}: Comparing the category abstract representations between humans and LMMs. (c) \textbf{Category Building}: Examining LMMs' categorization ability on abstract unseen categories.}
\label{fig:progressive-evaluation-tasks}
\vspace{-0.8em}
\end{figure*}

\textbf{Pattern Perception: pre-CL evaluation.} When perceiving entities in the concrete space, patterns are the direct perceptual targets for humans and LMMs. The ability to accurately identify low-level patterns is a prerequisite for category learning (pre-CL). 
As shown in Fig.\ref{fig:progressive-evaluation-tasks} (a), we evaluate the ability of LMMs to recognize patterns in multiple dimensions, such as shape, material, color, etc., in a fully disentangled manner.

\textbf{Abstraction Alignment: post-CL evaluation.}
Alignment between the abstract spaces is one of the important topics in eXplainability AI (XAI) research \cite{alignment}.
We further explore whether LMMs' learned representations of category learning (post-CL) are aligned with human mental representations.
As shown in Fig.\ref{fig:progressive-evaluation-tasks} (b), 
LMMs are asked to recognize abstract visual stimuli agreed upon by humans and align them with the correct semantic labels.

\textbf{Category Building: full-chain evaluation.} As shown in Fig.\ref{fig:progressive-evaluation-tasks} (c), to examine the categorization capability of LMMs from learning to use, we define several groups of abstract unseen categories in ComBo, requiring LMMs to observe exemplar objects, induce the definitions of two categories and classification boundaries, and finally classify new objects.
This task is considered complex as it involves both perception and cognition, closely resembling many laboratory tasks designed for human participants.


\begin{figure*}[t]
\begin{center}
\includegraphics[width=1.0\linewidth]{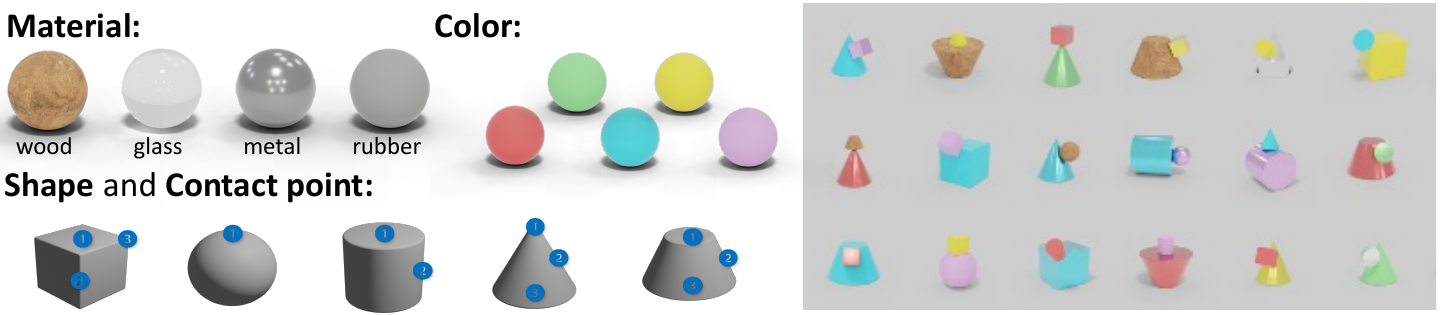}
\end{center}
\caption{Overview of \textbf{Com}posite \textbf{B}l\textbf{o}cks (ComBo) dataset: exemplar images and attributes. Each object can be represented by four-dimension fully-disentangled attributes as shape, color, material, and contact point between the primary primitive and the secondary primitive.}
\label{fig:dataset}
\vspace{-1.8em}
\end{figure*}

\vspace{-0.4em}
\section{The ComBo Benchmark}
\label{sec:combo}

\subsection{Overview of ComBo}

We construct a large-scale repository of \textbf{Com}posite \textbf{B}l\textbf{o}cks  for categorization (ComBo),
where each object within the dataset is composed of two geometric primitives, named primary primitive and secondary primitive according to the size of the primitives.
The primary and secondary primitives are contacted through a contact point on the primary primitive.
The optional shapes of the primitives and the optional contact points on the primary primitive are all displayed in Fig.\ref{fig:dataset}.
To enhance the visual diversity of ComBo, four different materials are assigned to the primitives. Additionally, the rubber and metal materials are further differentiated by five colors.

By enumerating all the values across the four disentangled dimensions of shape, material, color, and contact point, a total of 9,504 objects can be obtained, with each pair of objects differing in at least one dimension.
We utilize a ray tracing based rendering engine \cite{blender} to render each composite object from 20 random viewpoints, culminating in the 190,080 images in ComBo, inspired by CLEVR \cite{johnson2017clevr}.
More details are shown in supplementary materials.

The benchmark content and evaluation results are publicly available at: \url{https://fubin29.github.io/Blocks-as-Probes/}.

\vspace{-0.4em}
\subsection{Tasks}

\begin{figure*}[t]
\begin{center}
\includegraphics[width=1.0\linewidth]{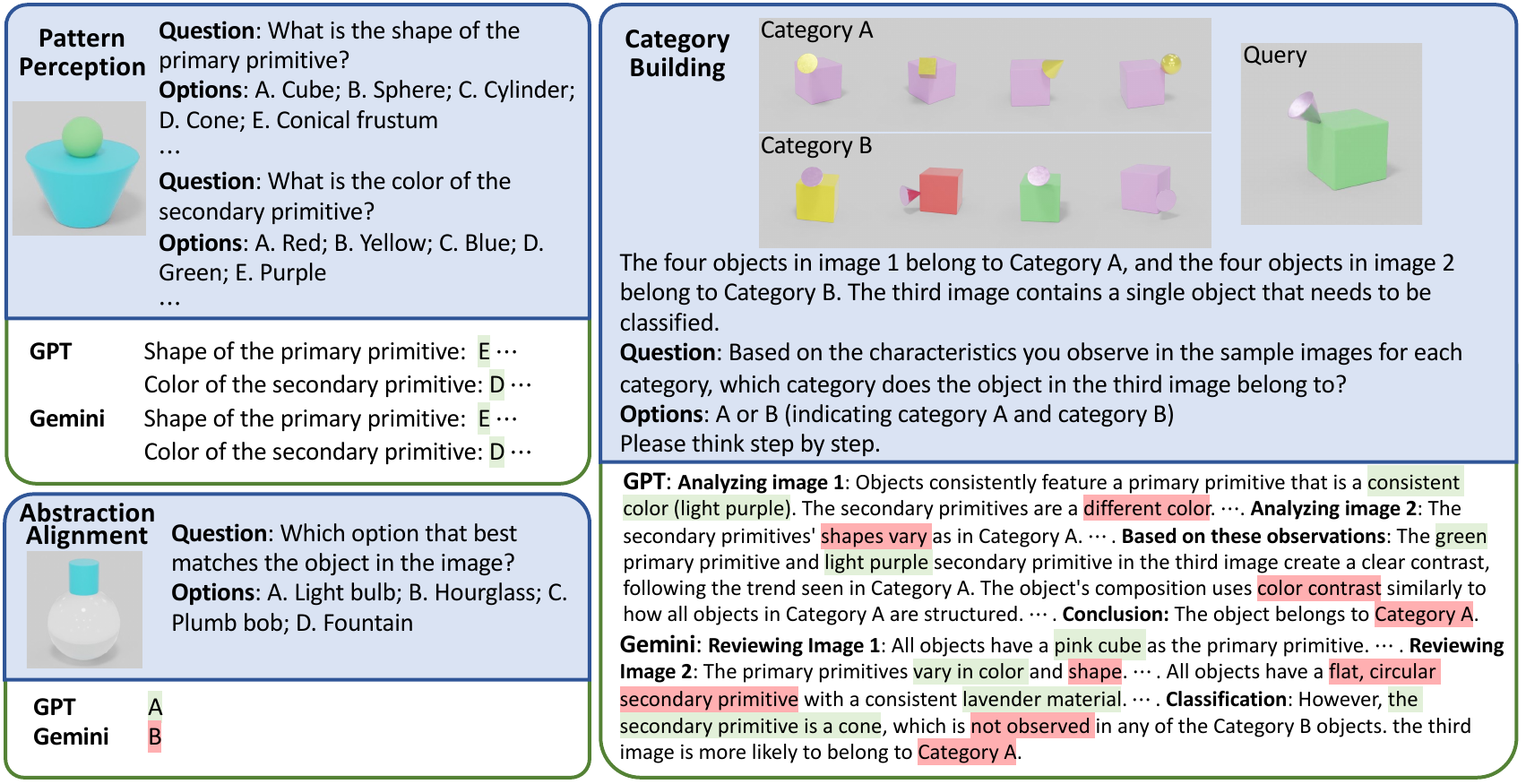}
\end{center}
\caption{Examples of the QA pairs for three evaluation tasks. Due to space constraints, prompts and answers are abbreviated. Refer to supplementary materials for details.}
\label{fig:question&ans}
\end{figure*}

As mentioned in Sec.\ref{sec:design-philosophy}, to evaluate LMMs' categorization capability, we start with three tasks: \textbf{Pattern Perception}, \textbf{Abstraction Alignment}, and \textbf{Category Building}.
Examples for three tasks and the corresponding answers by LMMs are illustrated in Fig.\ref{fig:question&ans}, and some statistics about the benchmark are shown in Tab.\ref{tab:task-details}.


\begin{table}[t]\small
    \begin{center}
    \begin{tabular}{lcccc}
    \toprule
    \textbf{Task Name} & \textbf{Evaluation Goal} & \textbf{\# Questions} & \textbf{Question Type} & \textbf{Metric} \\
    \midrule
    Pattern Perception  & pre-CL & 5000 & multiple-choice & accuracy \\
    \midrule
    Abstraction Alignment & post-CL & 240 & multiple-choice & accuracy \\
    \midrule
    \multirow{2}*{Category Building} & \multirow{2}*{full-chain} & 400 & multiple-choice & accuracy \\
	~ & ~ & 400 & CoT & manual scoring \\
    \bottomrule
    \end{tabular}
    \end{center}
    \caption{Summary of three evaluation tasks. For multiple-choice questions, we use the accuracy of correct options as evaluation metric. For answers by Chain-of-Thought (CoT), they are manually scored by experts to obtain accuracy, process score, and error attribution.}
    \label{tab:task-details}
    \vspace{-0.8em}
\end{table}


\textbf{Pattern Perception.}
We randomly select 5,000 objects from ComBo as evaluation subjects, and sample one rendered image for each object.
Participants are required to sequentially answer seven questions about the low-level patterns present in the object, as shown in Tab.\ref{tab:entity-perception}.
All questions are multiple-choice, and a brief description of ComBo along with all the options are provided.

\textbf{Abstraction Alignment.}
In this task, we invite cognitive science experts to select appropriate natural categories that can be abstracted by our ComBo objects.
Following a filtering and voting process, a consensus is reached on 24 categories, which include image samples and category labels, for evaluation purposes.
Subsequently, we generate two types of multiple-choice questions, each comprising 120 questions, by incorporating distractors among the matched abstract objects and category labels.
\textit{Img2Text} requires the participants to choose the label that best matches the given image out of four category labels.
\textit{Text2Img} asks the participants to select the image that most resembles the given category from four image options.
The detailed process of question generation is described in supplementary materials, and the validity of the questions is verified by user study discussed in Sec.\ref{sec:results}.


\textbf{Category Building.} 
In this task, we require participants to simulate the category formation in human consensus by constructing abstract categories of different granularities, and then classify the test samples.
Abstract categories are groups of object clusters defined based on rules, where all composite objects in an abstract category have the same constraint (e.g., ``with a red cube as the primary primitive'').
We present multiple samples from two abstract categories to the participants, requiring them to observe and summarize the rules for building both abstract categories.
Furthermore, we randomly show test samples belonging to the two categories to the participants multiple times.
Participants should be able to classify all samples correctly when they understand the categories.
We also design a similarity measurement method to calculate the evaluation task difficulty for classifying composite objects within abstract categories.
See supplementary materials for more details.


\section{Experiments}
\label{sec:exp}
\subsection{Evaluation Settings}

In this study, we select the mainstream closed-source implementations of current LMMs (GPT-4V \cite{gpt4vision, yang2023dawn}, Gemini-1.5-Pro \cite{reid2024gemini}) and open-source implementations (LLaVA-v1.5-13B \cite{liu2023improved}, Qwen-VL-Chat \cite{bai2023qwen}) as the subjects of our analyses.
To evaluate GPT-4V and Gemini, we utilize their official APIs. For LLaVA and Qwen, we conduct local tests using a single NVIDIA A40 GPU.
Additionally, we include other implementations as comparative references in different experiments, such as representation classification based on CLIP \cite{radford2021learning} pre-trained models, evaluations from human users, etc.
More details and examples about image input and prompt are presented in supplementary materials.



\subsection{Evaluation Results}
\label{sec:results}

\textbf{Pattern Perception.}
Tab.\ref{tab:entity-perception} demonstrates the pattern perception capability of various LMMs, without any fine-tuning or in-context prompting. Gemini and GPT-4V exhibit significantly stronger low-level pattern recognition and instruction-following capability, compared to open-source LMMs (even in simpler separate questions). Notably, Gemini and GPT-4V generally achieve the accuracy of larger than 90\% in recognizing the primary primitive’s shape and colors of both primitives. We also find that the recognition of patterns in smaller secondary primitives presents greater challenges, resulting in performance declines across all the LMMs. Overall, Gemini achieves the best results in all metrics, and especially excels in predicting the contact points, indicating its advanced spatial perception capability.

Considering the domain transfer challenges posed by the ComBo dataset, we conduct additional in-context learning experiments \cite{brown2020language, alayrac2022flamingo, tsimpoukelli2021multimodal} on GPT-4V and Gemini, focusing on contact points and materials of both primitives. As shown in Fig.\ref{fig:exp1-ICL}, the results indicate that GPT-4V can significantly improve its performance through in-context learning. We speculate that Gemini's performance advantage over GPT-4V might stem from Gemini's exposure to similar block data during training and specialized training on spatial relationships.


\begin{table}[t]\small
    \begin{center}
    \begin{tabular}{lccccccc}
    \toprule
    & \multicolumn{3}{c}{Primary Primitive} & \multicolumn{3}{c}{Secondary Primitive} & \multirow{2}*{Contact Point} \\
    \cmidrule(r){2-4}  \cmidrule(r){5-7}
    & Shape & Material & Color & Shape & Material & Color & ~ \\
    \midrule
    LLaVA   & 55.7 & 50.7 & 40.6 & 48.6 & 38.1 & 29.5 & 47.2 \\ 
    Qwen    & 70.7 & 70.2 & 65.8 & 37.2 & 43.1 & 25.4 & 43.8 \\
    \midrule
    GPT-4V  & 89.4 & 75.8 & 94.8 & 64.6 & 68.4 & 87.9 & 43.7\\
    Gemini  & \textbf{95.9} & \textbf{96.0} & \textbf{99.5} & \textbf{79.6} & \textbf{86.6} & \textbf{94.8} & \textbf{64.9} \\
    \bottomrule
    \end{tabular}
    \end{center}
    \caption{Pattern Perception Results (\%). For GPT-4V and Gemini, all seven low-level patterns are queried simultaneously in a single question, whereas each pattern is addressed in a separate question for LLaVA and Qwen.}
    \label{tab:entity-perception}
    \vspace{-0.8em}
\end{table}

\textbf{Abstraction Alignment.}
We invite 20 human participants to complete the user study to validate the reasonableness of our questions. The results indicate that the abstract objects in the images bear a good resemblance to the mental representation of natural categories held by humans.
Tab.\ref{tab:abstraction-alignment} presents the alignment between the category concepts learned by LMMs and human mental representations for natural categories. For \textit{Img2Text}, which involves matching one image to four semantic labels, all the LMMs exhibit similar performance with an accuracy rate of around 50\%. For \textit{Text2Img}, matching one semantic label to four images, GPT-4V and Gemini outperform the open-source LMMs. However, considering both types of questions, the abstract reasoning ability of LMMs still falls short of humans. 

Additionally, two conclusions can be drawn from the experimental results: (1) The poor performance of open-source LMMs on \textit{Text2Img} stems from the need to reason based on multiple input images, a relative weakness for these models compared to others like GPT-4V. (2) \textit{Img2Text} is harder than \textit{Text2Img}. In the semantic space, a single image corresponds to a relatively definite feature representation, while the feature representation for a label is actually the centroid of a group of similar images’ features. Therefore, matching images to multiple semantic labels in the question introduces greater uncertainty.

\begin{table}[t]\scriptsize
\begin{tabular}{cc}
    \begin{minipage}{.4\linewidth}
        \begin{center}
        \includegraphics[width=1\linewidth,height=0.6\linewidth]{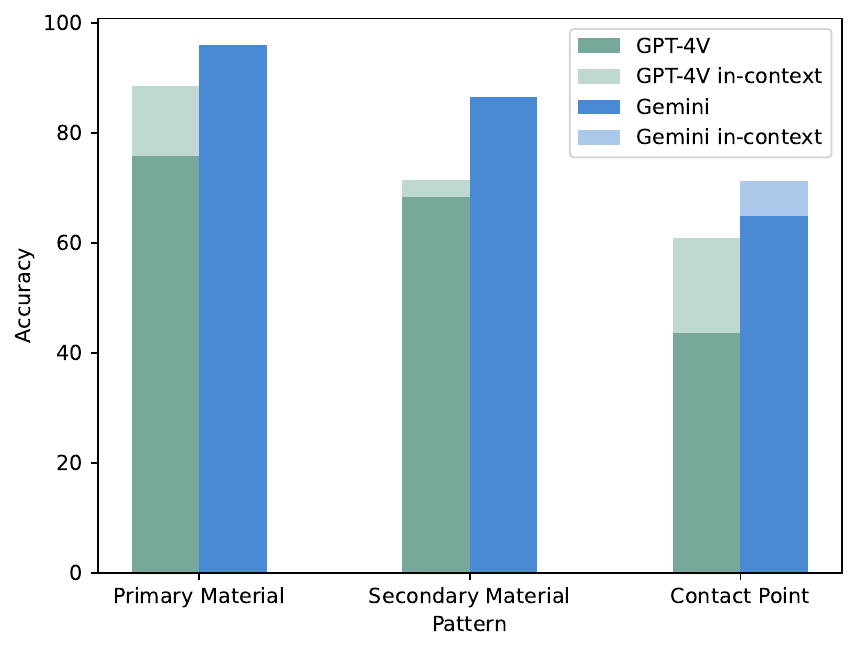}
        \end{center}
        \captionof{figure}{Comparison of Accuracy (\%) of Pattern Perception Task: GPT-4V and Gemini with and without in-context examples.}
        \label{fig:exp1-ICL}
    \end{minipage} &
    \begin{minipage}{.5\linewidth}
        \begin{center}
        \begin{tabular}{lccc}
            \toprule
            & \textit{Img2Text} & \textit{Text2Img} & Overall \\ \midrule
            CLIP    & 47.5  & 75.8  & 61.7 \\ \midrule
            LLaVA   & \textbf{55.0}  & 29.2  & 42.1 \\
            Qwen    & 48.3  & 32.5  & 40.4 \\ \midrule
            GPT-4V  & \textbf{55.0}  & \textbf{76.7}  & \textbf{65.9} \\
            Gemini  & 52.5  & 74.2  & 63.4 \\ \midrule
            Human   & \textit{80.8 $\pm$ 9.5}  & \textit{94.4 $\pm$ 5.2}  & \textit{87.6 $\pm$ 10.2} \\ \bottomrule
        \end{tabular}
        \end{center}
        \caption{Abstraction Alignment Results (\%). Specifically, by calculating the similarity between CLIP features of the query and the options, we provide CLIP's alignment results.}
        \label{tab:abstraction-alignment}
    \end{minipage} 
\end{tabular}
\vspace{-0.8em}
\end{table}


\textbf{Category Building.}
In this task, we invite 8 human participants to assess whether the proposed similarity measurement method correlates with human cognition and whether it is applicable for evaluating the classification difficulty of abstract categories.
Based on this similarity measurement, we design four difficulty levels: easy, medium, hard, and expert. We also invite another 23 human participants to complete this task, providing a human reference score (the red line in Fig.\ref{fig:category-building}) for comparative analysis.
Fig.\ref{fig:category-building} illustrates the complete categorization capability of different LMMs across varying difficulties.\footnote{Qwen failed to produce a valid output in this task, hence there is no corresponding entry in Fig.\ref{fig:category-building}}
As the difficulty increases, represented by the diminishing differences between two categories, the challenge for LMMs to construct accurate category representations and classify query objects also escalates, and all LMMs demonstrate varying degrees of performance decline.
In contrast, human participants maintain high classification accuracy across different difficulty levels. Even in the expert level, human participants can perceive the increase in difficulty but still manage to cope effortlessly. 
However, when the differences between two categories become minimal, and they must rely on the shape of secondary primitives or the combination of two primitives for distinction, the weaknesses of LMMs in pattern perception are further magnified in this experiment.

We further employ the Chain-of-Thought (CoT) approach \cite{wei2022chain, brown2020language}.
Both GPT-4V and Gemini possess robust reasoning ability; they can effectively perform category building and application when objects are accurately perceived. The performance differences displayed by the two models in more difficult questions primarily stem from Gemini's advantage in low-level pattern recognition. From Fig.\ref{fig:category-building}, requiring LMMs to explicitly output their decision-making processes in CoT format indeed enhances their performance in this task.
Moreover, CoT technology helps us understand the causes of misclassification, as shown in Fig.\ref{fig:category-building-cot}.

\begin{table}[t]\small
\begin{tabular}{cc}
    \begin{minipage}{.55\linewidth}
        \begin{center}
        \includegraphics[width=1\linewidth,height=0.6\linewidth]{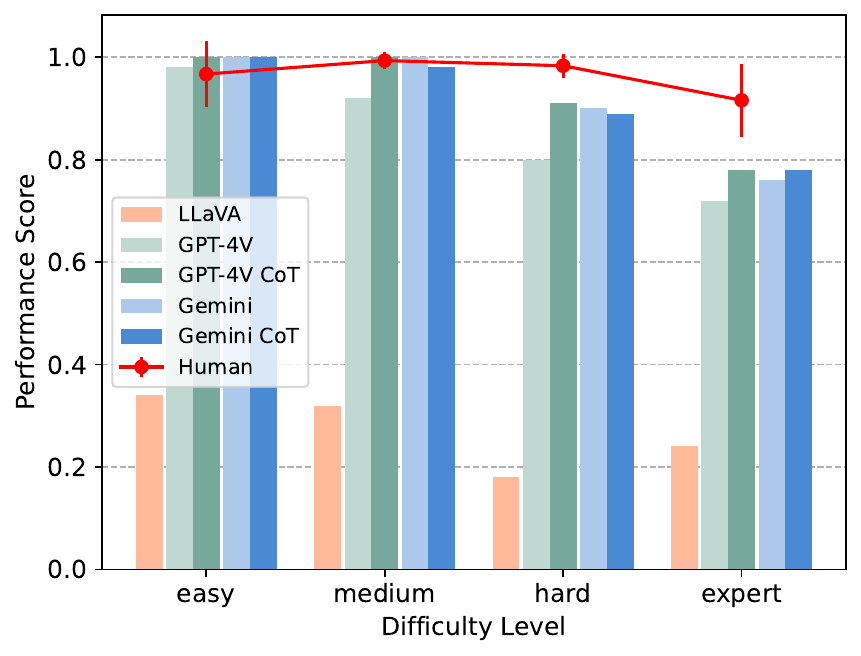}
        \end{center}
        \captionof{figure}{Category Building Results. GPT-4V and Gemini both demonstrate strong performance on this task, with GPT-4V achieving greater improvements when employing the CoT technique.}
        \label{fig:category-building}
    \end{minipage} &
    \begin{minipage}{.35\linewidth}
        \begin{center}
        \includegraphics[width=1\linewidth,height=0.85\linewidth]{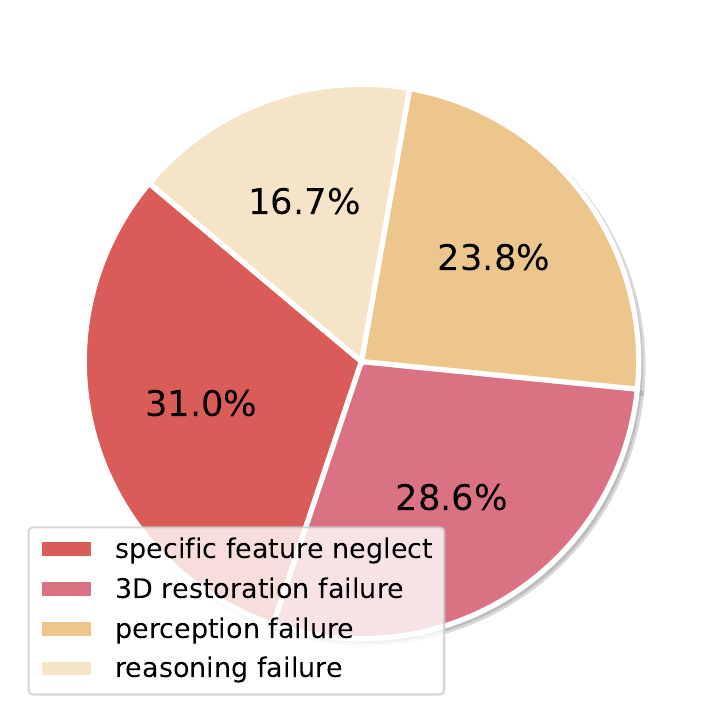}
        \end{center}
        \captionof{figure}{Error Analysis of LMMs Using CoT Technology: distribution of error types in LMMs' decision-making processes.}
        \label{fig:category-building-cot}
    \end{minipage} 
\end{tabular}
\vspace{-0.8em}
\end{table}

\subsection{Discussion}
\label{sec:disussion}

\subsubsection{Prompt Design and Instruction Following}
During the evaluation process, we find significant differences in the instruction following \cite{ouyang2022training} capability among different LMMs. GPT-4V and Gemini can understand the most complete description of the questions, and can also receive multiple image inputs, allowing them to handle more complex problems that require reasoning between multiple images.
In pattern perception evaluation, querying different patterns separately can improve the performance of LLaVA. Qwen's capability to follow complex instructions is slightly inferior to other LMMs, requiring more adjustments to the form of the questions.

\subsubsection{Task Difficulty}

In this section, we discuss the \textbf{difficulty} of our benchmark.
Although ComBo, as a dataset rendered based on geometric primitives, defines problems that are formally similar to abstract reasoning problems \cite{fu2023challenger} in other benchmarks, 
it is much less difficult than those benchmarks in terms of logical reasoning.
To verify this, we use some smaller, commonly used computer vision models
to complete the same tasks.
By retrieval, models pre-trained on ImageNet-1k \cite{ILSVRC15} can achieve similar performance to GPT-4V, and almost are able to completely solve the task after fine-tuning.
See supplementary materials for more details.
The fine-tuned small model is capable of completing the task,
implying that it does not necessitate a complex reasoning process.
However, current LMMs still show a significant gap compared to humans in such simple visual tasks,
indicating that LMMs are \textbf{far from being able to claim} that the fundamental tasks of visual categorization \textbf{have been completely solved}.

\subsubsection{Limitation and Future Work}
We believe that evaluating both the lower and upper bounds of LMMs' capabilities is equally important. Compared to other comprehensive evaluation benchmarks, our evaluation benchmark leans more towards the in-depth evaluation of \textbf{categorization capability}, which is considered one of the most fundamental visual cognitive abilities.
Notably, we utilize brand new synthetic data rather than real 2D images to completely \textbf{prevent data leakage} and facilitate \textbf{decoupled}, \textbf{controllable} evaluation.
However, it is still necessary to use more complex, even real images to further evaluate the capabilities of LMMs in real-world application scenarios.
In future work, we will involve more complex composite objects and controllable 3D models, and incorporate a wider array of cognitive tasks to 
further explore the current conclusion.
Moreover, we are committed to developing new methodologies and datasets that enhance LMMs' performance in perceiving spatial details, reasoning about abstract concepts, and learning new categories.
We believe these efforts will enhance the applicability and reliability of LMMs in various high-level tasks.

\section{Related Work}
\label{sec:rw}

\textbf{Large Multimodal Models.} LMMs \cite{yin2023survey, awadalla2023openflamingo, dai2024instructblip} integrate visual \cite{liu2024visual} or other modalities \cite{wu2023next, guo2023point} into Large Language Models (LLMs), enabling them to handle a variety of multimodal tasks.
High-performance and closed-source LMMs like PaLM-E \cite{driess2023palm}, GPT-4V \cite{yang2023dawn}, and Gemini \cite{team2023gemini} represent a critical branch of development. These models benefit from substantial investments in proprietary datasets and computing resources, achieving superior performance across a range of complex tasks. Another branch consists of open-source models such as LLaMA-Adapter \cite{zhang2023llama}, LLaVA \cite{liu2024visual}, MiniGPT-4 \cite{zhu2023minigpt}, Otter \cite{li2023mimicit}, and Qwen \cite{bai2023qwen}. These LMMs are typically developed by modularly integrating other modalities into open-source LLMs \cite{touvron2023llama}. Both branches have demonstrated strong capability in various applications, such as medical image understanding \cite{li2024llava, moor2023med} and embodied agents \cite{wang2024mobile, yang2023octopus}. Consequently, we select two models from each branch for evaluation in our study.

\noindent\textbf{LMM Benchmarks.} Due to the more generalized multimodal perception and reasoning capability of LMMs, traditional vision-language benchmarks are inadequate for providing a comprehensive and sufficient evaluation. Consequently, recent developments in the evaluation of LMMs have primarily focused on several key aspects \cite{yin2023survey}: (1) addressing specific common issues such as visual shortcomings \cite{tong2024eyes} and hallucinations \cite{cui2023holistic, liu2023hallusionbench}; (2) comprehensive benchmarks that entail complex tasks and diverse capability \cite{yin2024lamm, xu2023lvlm, li2023seed, liu2023mmbench, yu2023mm}; (3) expert-level domain knowledge and advanced reasoning \cite{yue2023mmmu}. In contrast to these evaluation efforts, our study concentrates on assessing the fundamental categorization ability of LMMs.

\section{Conclusion}

In this work, we introduce the ComBo benchmark, focusing on evaluating the categorization capability of Large Multimodal Models (LMMs). Inspired by research on categorization in cognitive science, we design three evaluation tasks from different perspectives, comprehensively assessing the LMMs' ability in pattern perception, abstract concept alignment, and generalization of categorization. The evaluation results reveal that LMMs still exhibit deficiencies in spatial detail perception, abstract concept reasoning, and learning of new categories. Although in-context learning or Chain-of-Thought (CoT) techniques can further improve the performance of LMMs, there remains a gap compared to human categorization capability, providing recommendations for future improvements in LMMs.

\newpage

\section*{Acknowledgement}
This work is partially supported by National Key R\&D Program of China No. 2021ZD0111901, 2023YFF1105104, and Natural Science Foundation of China under contract No. U21B2025.
Bin Fu and Qiyang Wan contributed equally to this work. Bin Fu was primarily responsible for the construction of the dataset and also participated in the design and implementation of the evaluation benchmark. Qiyang Wan mainly provided theoretical formulation in categorization within cognitive science and similarly contributed to the design and implementation of the evaluation benchmark.

\newpage

\appendix

\section{Introduction}
In this supplementary material, we elaborate on 
more details including dataset samples (\blue{Sec.3.1 in the main paper}\footnote{For better understanding, the mentioned figures, tables, sections in the main paper are denoted in blue.}) 
and evaluation tasks (\blue{Sec.3.2 in the main paper}) of our ComBo benchmark in Sec.\ref{sec:combo}, 
more details on evaluation settings including configuration of LMMs, prompts (\blue{Sec.4.1 in the main paper}), and user studies in Sec.\ref{sec:eval-setting}, 
more supplementary examples for experimental results (\blue{Sec.4.2 in the main paper}) in Sec.\ref{sec:result}, 
and extra discussion on task difficulty (\blue{Sec.4.3 in the main paper}) in Sec.\ref{sec:discussion}.
As mentioned in the main paper, the full ComBo benchmark and all evaluation results will be released to the public upon acceptance.


\section{The ComBo Benchmark}
\label{sec:combo}
In this section, we first provide a detailed introduction to the construction process of composite objects and the dataset generation process (Sec.\ref{sec:2-1}). Subsequently, we elaborate on the specific details of three evaluation tasks (Sec.\ref{sec:2-2}-\ref{sec:2-4}).

\subsection{Object Assembly and Image Rendering}
\label{sec:2-1}

\begin{figure*}[htb]
\begin{center}
\includegraphics[width=0.9\linewidth]{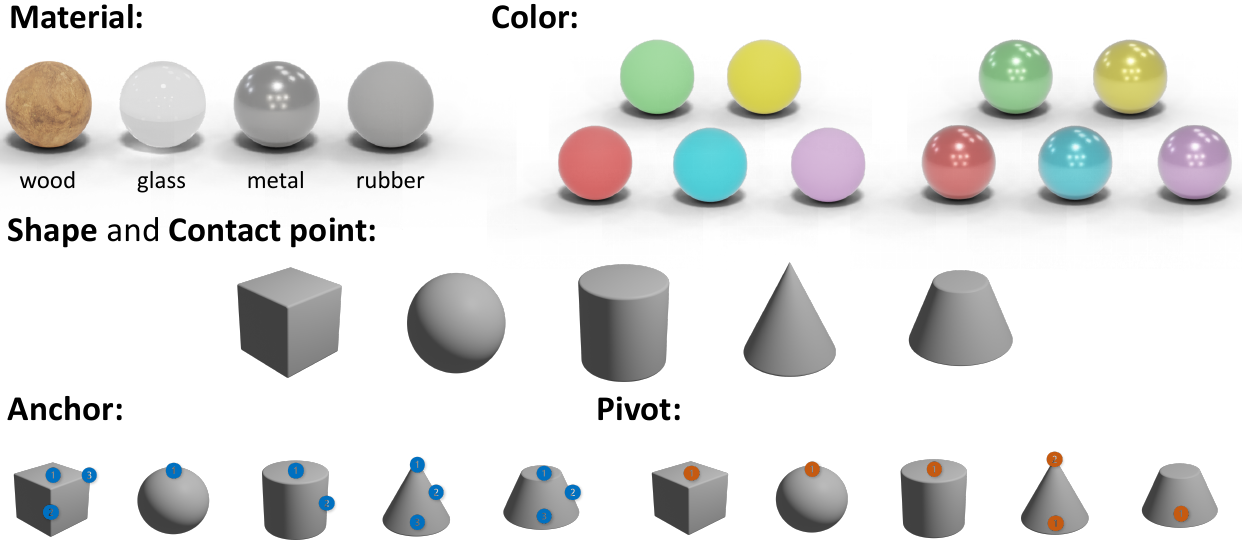}
\end{center}
\caption{Types of patterns in ComBo.}
\label{fig:anchor-pivot}
\end{figure*}

Each object within \textbf{Com}posite \textbf{B}l\textbf{o}cks (ComBo) is composed of two geometric primitives. These primitives vary in two distinct sizes: the larger one is referred to as the primary primitive, and the smaller one is termed the secondary primitive.
The geometric primitives are available in five shapes: cube, sphere, cylinder, cone, and conical frustum.
During the assembly process, both the primary and secondary primitives select a contact point, denoted as the anchor and pivot, respectively, as shown in Fig.\ref{fig:anchor-pivot}.
The secondary primitive's pivot is then attached to the primary primitive's anchor at a predefined angle.

It is noted that among secondary primitives, only cone has two selectable pivot options.
Therefore, for other shapes, when examining the Large Multimodal Models' (LMMs) perception of pivots through multiple-choice questions, there is only one possible answer, which solely assesses whether the LMMs can legally produce the unique correct answer.
Although we have actually evaluated the LMMs' perception of pivots, due to the limited practical significance of this metric and space limitations, we did not discuss pivots and related experiments in the main paper.
We simplified the assembly process of primitives in the main paper to state that the primary and secondary primitives are connected through a contact point on the primary primitive.
Here, the true meaning of the ``contact point'' is actually the anchor selected during the assembly process of the objects.

To enhance the visual diversity of ComBo, four different materials are assigned to the geometric primitives: rubber, metal, glass, and wood.
Additionally, the rubber and metal materials are further differentiated by five colors: red, yellow, blue, green, and purple.

\begin{figure*}[htb]
\begin{center}
\includegraphics[width=1.0\linewidth]{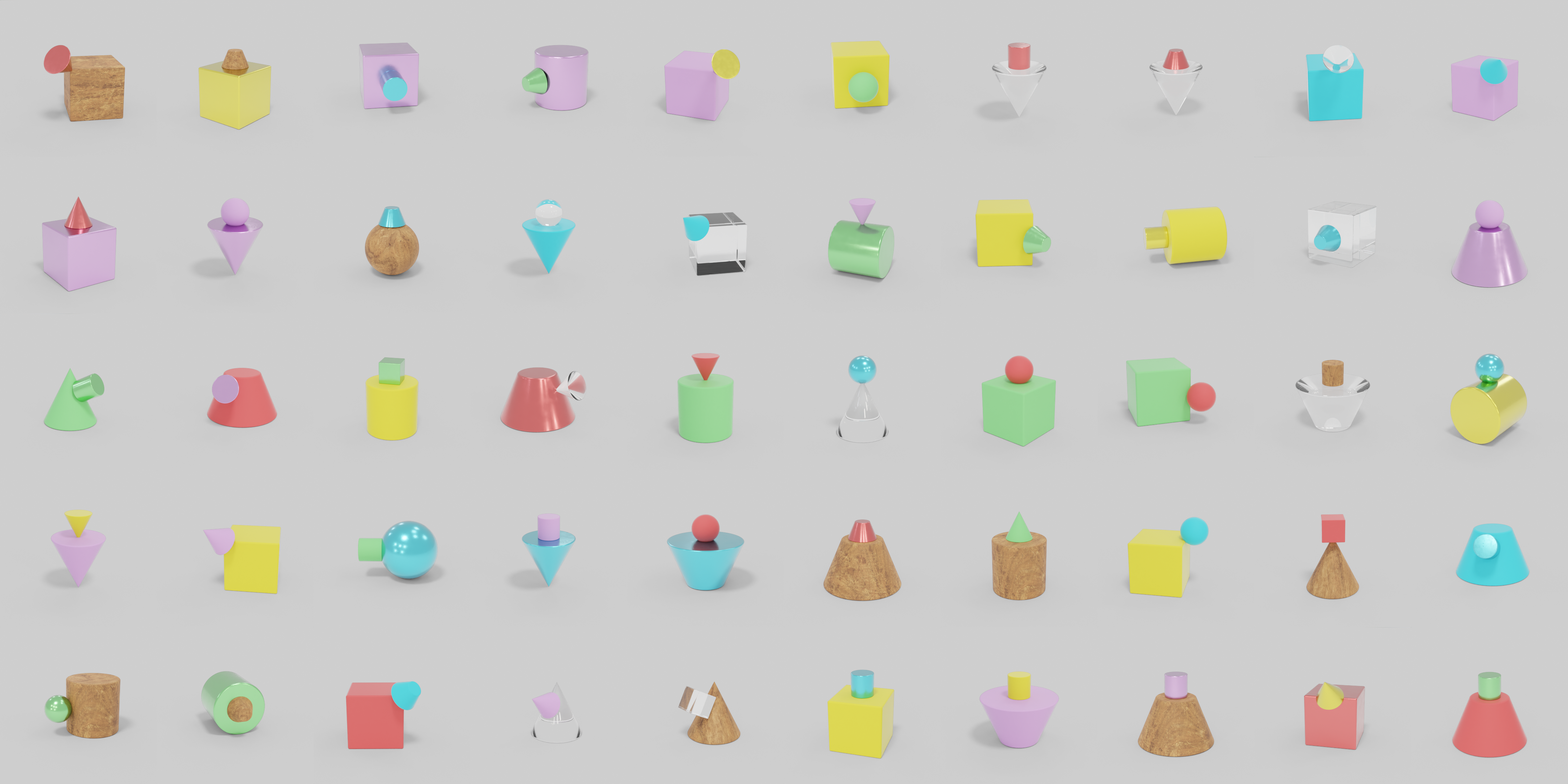}
\end{center}
\caption{More exemplar images in ComBo.}
\label{fig:gallery}
\end{figure*}

We utilize a ray tracing based rendering engine \cite{blender} to render each composite object, thereby obtaining more photorealistic images, inspired by CLEVR \cite{johnson2017clevr}.
The composite objects are positioned at the center of the scene, with appropriate environment lighting and several point lights. Each object is rendered from 20 random viewpoints, culminating in the 190,080 images in ComBo.
More composite objects are demonstrated in Fig.\ref{fig:gallery}.

\subsection{Pattern Perception}
\label{sec:2-2}
Pattern Perception is the most fundamental visual perception task, requiring subjects to answer a set of questions about composite objects, including shape, color, material, and contact point.
It is important to note that the contact point consists of the anchor on the primary primitive and the pivot on the secondary primitive, together forming the manner of object composition.
However, for most geometric primitives, the choice of pivot when acting as a secondary primitive is unique (with only \textit{cone} being the exception).
Therefore, when we discuss the contact point in the main paper, we are primarily referring to the anchor on the primary primitive.

\subsection{Abstraction Alignment}
\label{sec:2-3}

In this task, we invite cognitive science experts to select appropriate natural categories that can be abstracted by our ComBo objects.

\begin{figure*}[thb]
\begin{center}
\includegraphics[width=1.0\linewidth]{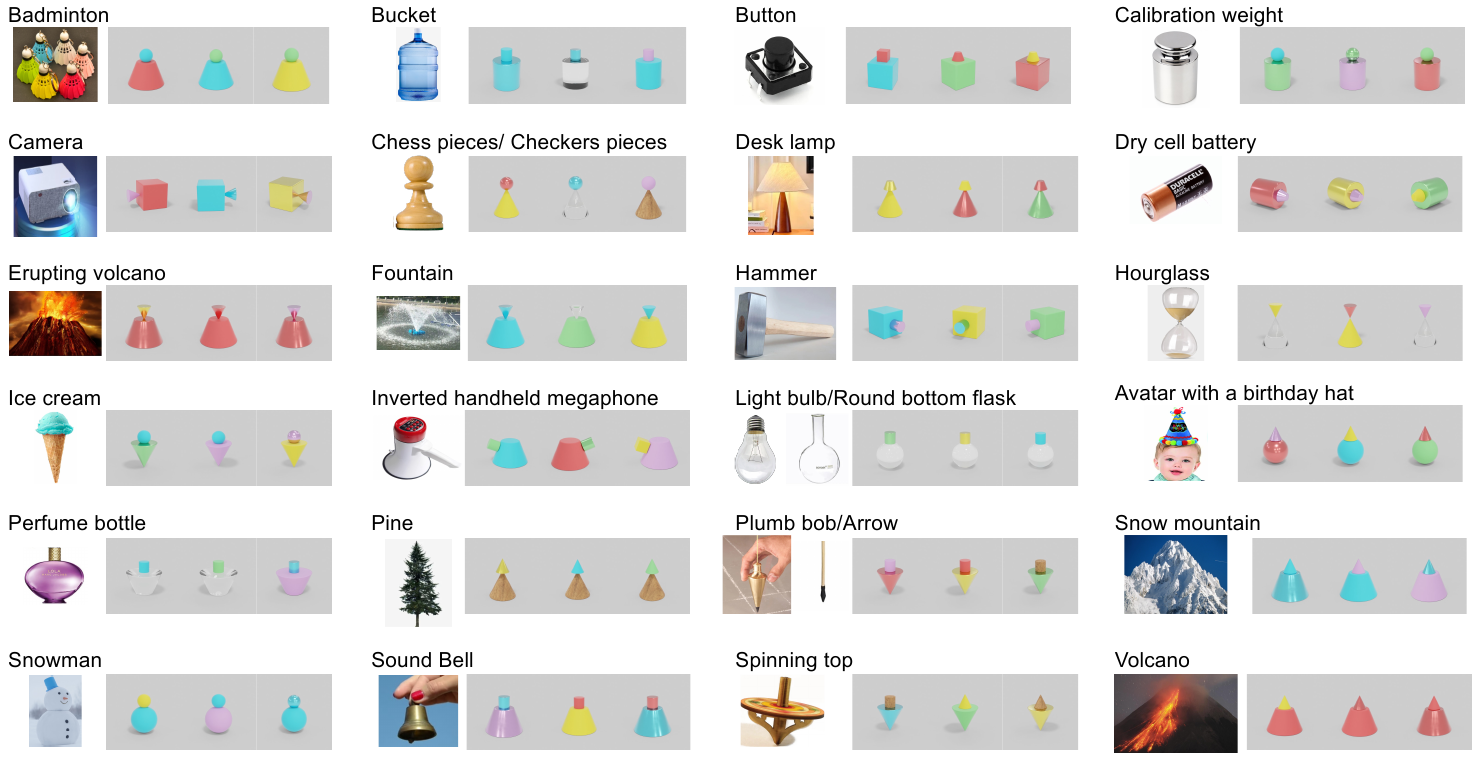}
\end{center}
\caption{Category labels and corresponding images selected in \textbf{Abstraction Alignment}.}
\label{fig:exp2-gallery1}
\end{figure*}

Specifically, the cognitive science experts select objects from the dataset that correspond significantly to the natural categories in human cognition, such as ``ice cream''.
After filtering and voting, a final consensus of 24 categories, including image samples and category labels, are selected for evaluating, with specific categories and corresponding images are listed in Fig.\ref{fig:exp2-gallery1}.

In this task, the color and material of objects are usually ignored, while shape plays a crucial role (according to the human cognition).
Additionally, although lighting and viewpoint are generally irrelevant, some specific categories are sampled under particular viewpoints if their pose greatly impacts category recognition.

\begin{figure*}[thb]
\begin{center}
\includegraphics[width=0.6\linewidth]{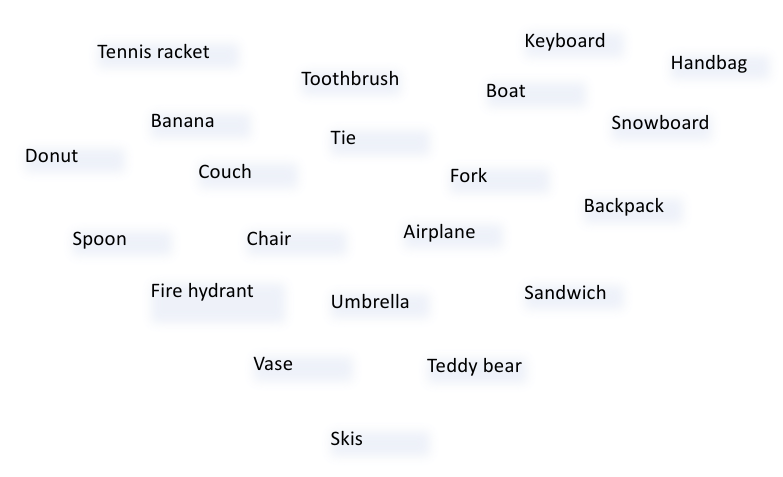}
\end{center}
\caption{Distractor categories from COCO.}
\label{fig:exp2-gallery2}
\end{figure*}

Participants are asked two types of multiple-choice questions.
\textit{Img2Text} requires the participants to choose the label that best matches the given image out of four category labels.
The four options include one correct answer and three distractors, which are selected from the remaining 23 categories and 21 categories introduced from the COCO dataset (shown in Fig.\ref{fig:exp2-gallery2}).
Since this task evaluates abstraction alignment capabilities of LMMs, we ask cognitive science experts to avoid choosing distractor categories that are very similar to the query image to prevent the problem from being overly difficult.
\textit{Text2Img} asks the participants to select the image that most resembles the given category from four image options.

\subsection{Category Building}
\label{sec:2-4}

As discussed in the main paper, in this task, we require  participants to simulate the category formation in human consensus by constructing abstract categories of different granularities, and then classify the test samples.
Abstract categories are groups of object clusters defined based on rules, where all composite objects in an abstract category have the same constraint (e.g., ``with a red cube as the primary primitive'').
The more constraints imposed, the fewer the dimensions of patterns that can be freely chosen, resulting in fewer composite objects included within an abstract category and a finer granularity of the category.

It is worth noting that the difficulty of the task is related to the selected abstract categories.
When the support samples from two categories are easily confused, summarizing the rules can become challenging.
To further model the task difficulty, we consider extracting visual features of all composite objects in the abstract categories using a general visual model (e.g., CLIP) to quantitatively characterize the classification difficulty based on the distribution distances of the samples.
Through cross-validation and user study, we finally adopt the solution of PCA dimensionality reduction, Gaussian parameter estimation, and Wasserstein distance as metrics.
Under this metric, the difficulty of distinguishing between two abstract categories is close to the difficulty in human cognition.

\section{Evaluation Settings}
\label{sec:eval-setting}

\subsection{Details about LMMs}
As mentioned in \blue{Sec.4.1 in the main paper}, we select the mainstream closed-source implementations of current LMMs (GPT-4V \cite{gpt4vision, yang2023dawn}, Gemini-1.5-Pro \cite{reid2024gemini}) and open-source implementations (LLaVA-v1.5-13B \cite{liu2023improved}, Qwen-VL-Chat \cite{bai2023qwen}) as the subjects of our study.

To evaluate GPT-4V and Gemini, we utilize their official APIs. For LLaVA-v1.5-13B and Qwen-VL-Chat, we conduct local tests using a single NVIDIA A40 GPU.
Considering the continuous updates to GPT-4V, we use different versions tailored to each task: for the pattern perception task, we use gpt-4-1106-vision-preview; for the abstraction alignment and category building tasks, we employ gpt-4-turbo-2024-04-09.
Specifically, we set the temperature of GPT-4V as zero in all tasks.
For all three evaluation tasks, we utilize the most recent version of Gemini, gemini-1.5-pro-latest, which was updated in April 2024.

LLaVA-v1.5-13B is an open-source chatbot trained by fine-tuning Vicuna v1.5 on GPT-generated multimodal instruction-following data.
Qwen-VL-Chat is also a vision-language chatbot, based on Qwen-VL-7B, with enhanced capabilities in following instructions.


\subsection{Prompts and Multiple Image Input}

\begin{figure*}[thb]
\begin{center}
\includegraphics[width=1.0\linewidth]{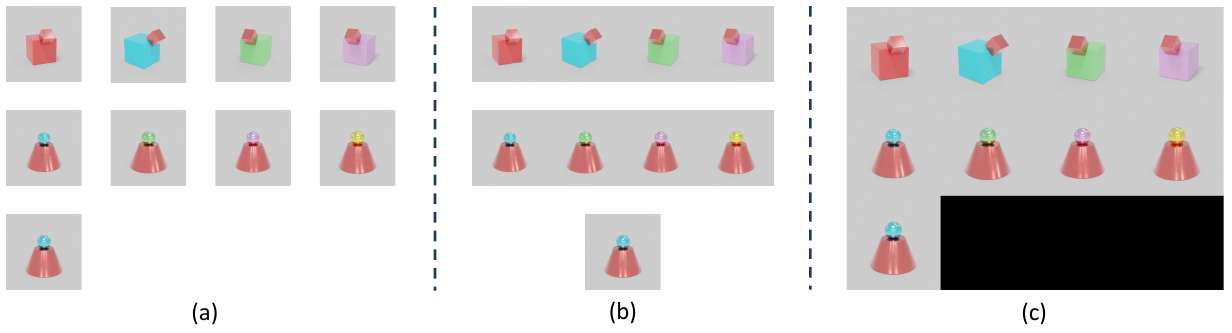}
\end{center}
\caption{Taking \textbf{Category Building} task as an example, we illustrate three methods of presenting images to LMMs. In method (a), nine images are passed to the LMMs as separate files. In method (b), four images belonging to the same abstract category are concatenated, reducing the number of images to simplify the reasoning process. In method (c), all images are combined into one large image, which is useful for LMMs that can only accept a single image input.}
\label{fig:multiple-image}
\end{figure*}

For all LMMs, our evaluation begins with the most comprehensive prompts. These prompts are intuitively designed to include system information about the ComBo benchmark, questions, multiple-choice options, and explicit instructions.
In cases where some LMMs are incapable of processing multiple images, we merge these images into a single, larger composite image and make corresponding modifications to the prompts, as shown in Fig.\ref{fig:multiple-image}.
For models with less robust instruction-following capability, excessively lengthy prompts and complex questions may hinder their ability to generate coherent responses.
Consequently, we iteratively shorten or modify the prompts to facilitate normal output.
These efforts are primarily aimed at ensuring that open-source LMMs can produce appropriate content, considering the different capabilities and preferences of various LMMs.
It is important to emphasize that all adjustments to the images and prompts are made to ensure a fair evaluation, tailoring only the format of the problem to align with the capability of each model.

\subsection{User Study}

We conducted multiple user studies to ensure the reasonableness of the questions and to provide human performance on different tasks for comparison with LMMs' capabilities.

In the \textbf{Abstraction Alignment} task, we invited 20 participants to complete the user study, with each participant answering 20 questions for each of the two question types.
After removing the highest and lowest total scores, we calculate the accuracy rates for each question type and the overall accuracy rate.
The results of the user study indicate that the questions constructed by the cognitive science experts align with human perception of natural categories.
Therefore, these data can be used to assess the \textbf{Abstraction Alignment} capabilities of LMMs compared to humans.

In the \textbf{Category Building} task, the user study was divided into two parts.
First, we invited 8 participants to rank the difficulty of classifying abstract categories.
Participants needed to complete 40 questions, each containing two options, with each option showing representative images of two abstract categories.
Participants were asked to determine which pair of the two abstract categories is more difficult to separate.
The aggregated answers were then compared with the Wasserstein distance between the two sets of categories, achieving a similarity of 95.4\%, which demonstrates that this metric aligns with human cognition and can be used to measure the classification difficulty of abstract categories.
Subsequently, we invited 23 participants to complete the \textbf{Category Building} task and recorded their scores across different task difficulties.

\begin{figure*}[htb]
\begin{center}
\includegraphics[width=1.0\linewidth]{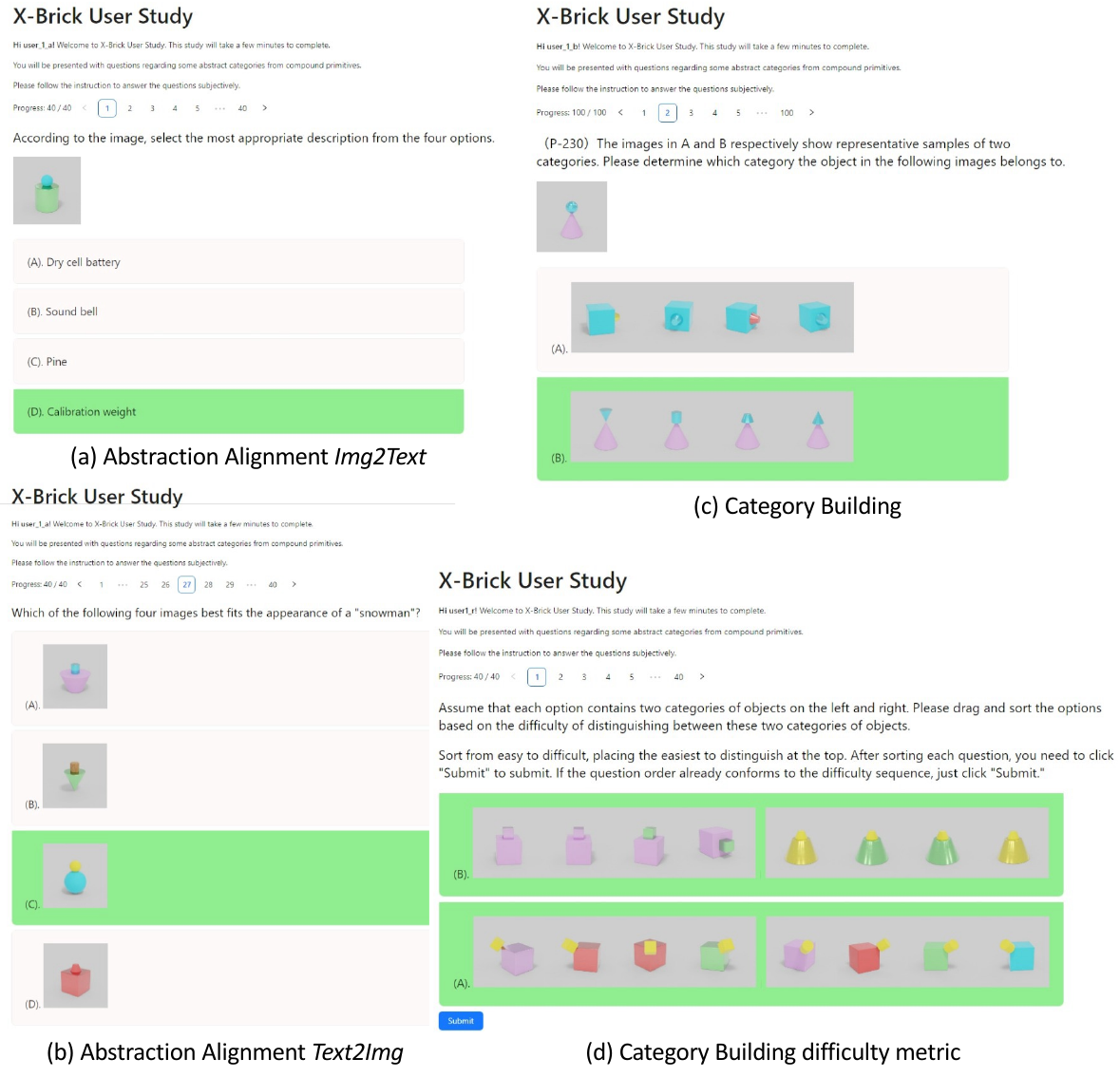}
\end{center}
\caption{Webpage screenshots of user studies.}
\label{fig:user-study}
\end{figure*}

The webpage used for completing the user studies is shown in Fig.\ref{fig:user-study}.

\section{Evaluation Results}
\label{sec:result}

In this section, we present the specific prompts used to evaluate LMMs across different tasks, as well as the responses from various LMMs.

As online LMMs continue to evolve and their outputs are non-deterministic, the actual output for a specific example cannot be guaranteed to be consistent with the results shown in the text.
However, the overall score is derived from a large number of questions, which is statistically significant.

\subsection{Pattern Perception}

To save costs and fully utilize the batch inference capabilities of LMMs, we adopt the method described in \cite{wu2023gpt4vis}, grouping multiple samples into one batch for a single query. In our experiments, we select the same batch size of 10 as recommended in \cite{wu2023gpt4vis}, which has almost no negative impact on the inference results.
In addition, based on the principle that an option should not be regarded as a test point when it is the only choice, in experiments concerning in-context learning where the anchor of the \textit{sphere} has only one option, we do not consider cases where the primary primitive is a \textit{sphere} in our experiments when evaluating the capability to recognize anchors.

\begin{figure*}[thb!p]
\begin{center}
\includegraphics[width=1.0\linewidth]{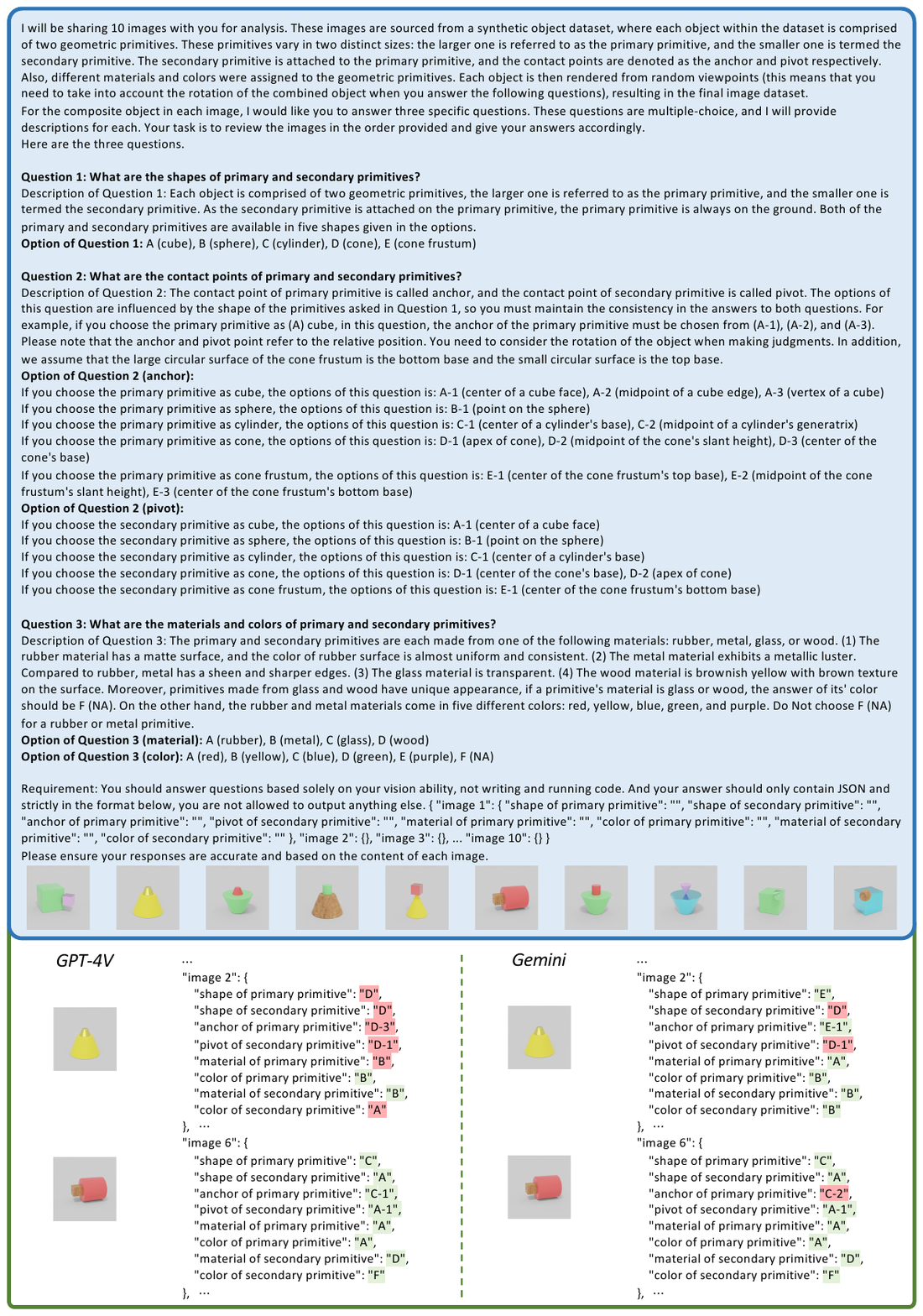}
\end{center}
\caption{Examples of the \textbf{Pattern Perception} task, using standard prompts, showcasing partial outputs from GPT-4V and Gemini.}
\label{fig:exp1-prompt-gpt}
\end{figure*}

\begin{figure*}[thb!p]
\begin{center}
\includegraphics[width=1.0\linewidth]{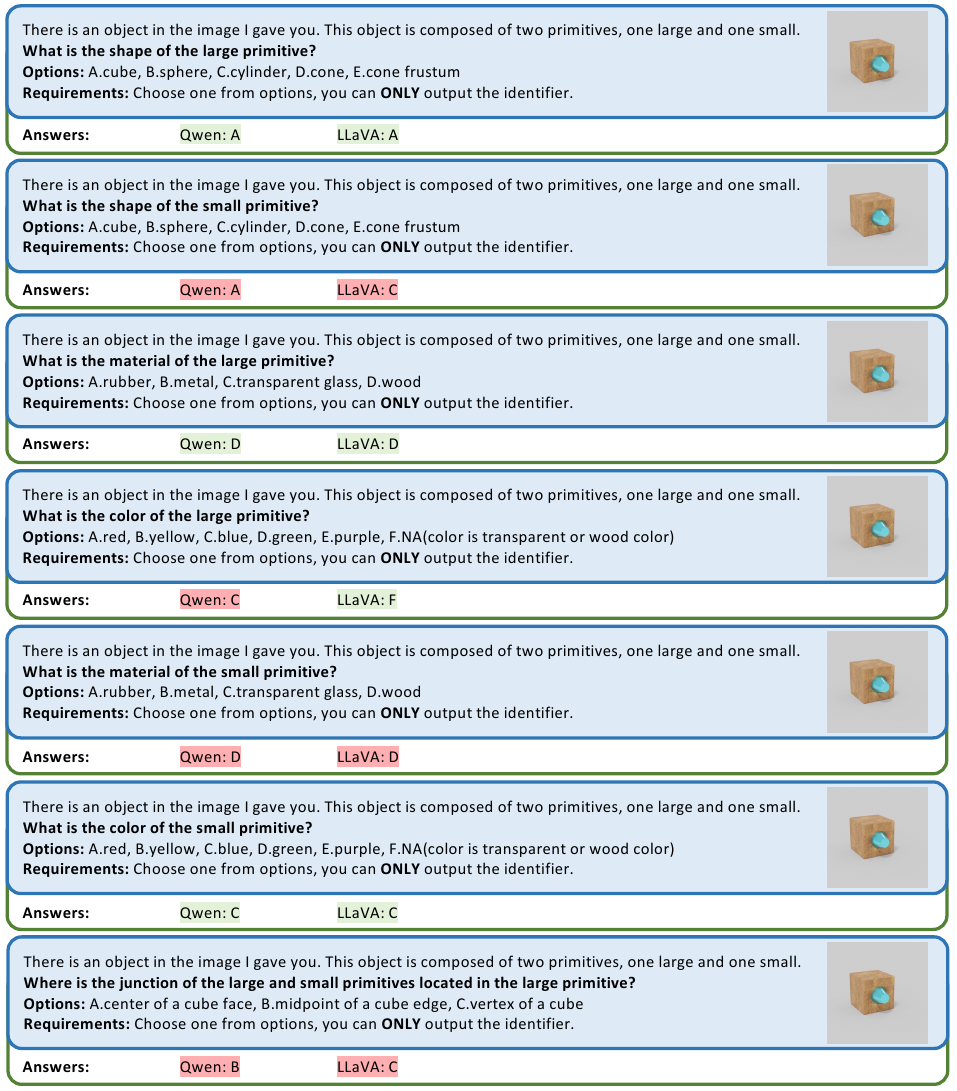}
\end{center}
\caption{An example of the \textbf{Pattern Perception} task, we use simpler prompts for LLaVA and Qwen.}
\label{fig:exp1-prompt-llava1}
\end{figure*}

\begin{figure*}[thb!p]
\begin{center}
\includegraphics[width=1.0\linewidth]{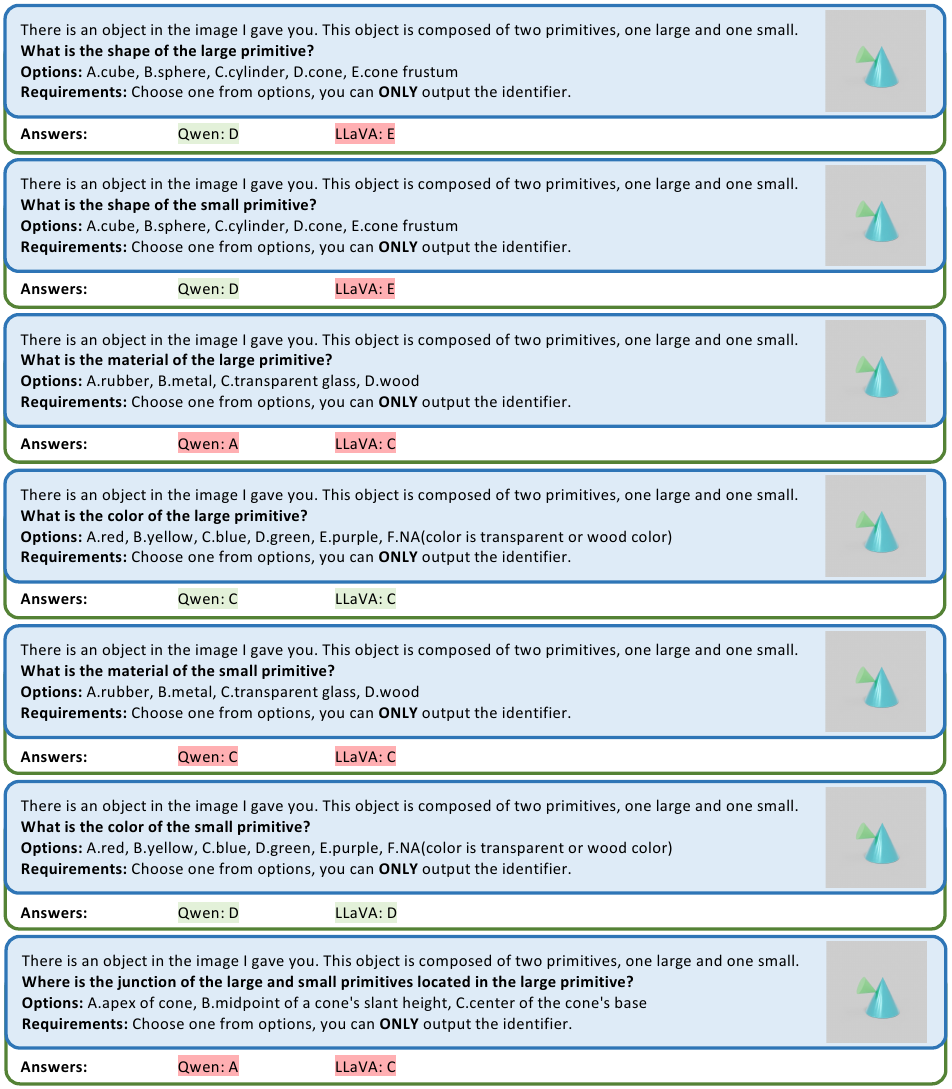}
\end{center}
\caption{An example of the \textbf{Pattern Perception} task, we use simpler prompts for LLaVA and Qwen.}
\label{fig:exp1-prompt-llava2}
\end{figure*}

\begin{figure*}[thb!p]
\begin{center}
\includegraphics[width=1.0\linewidth]{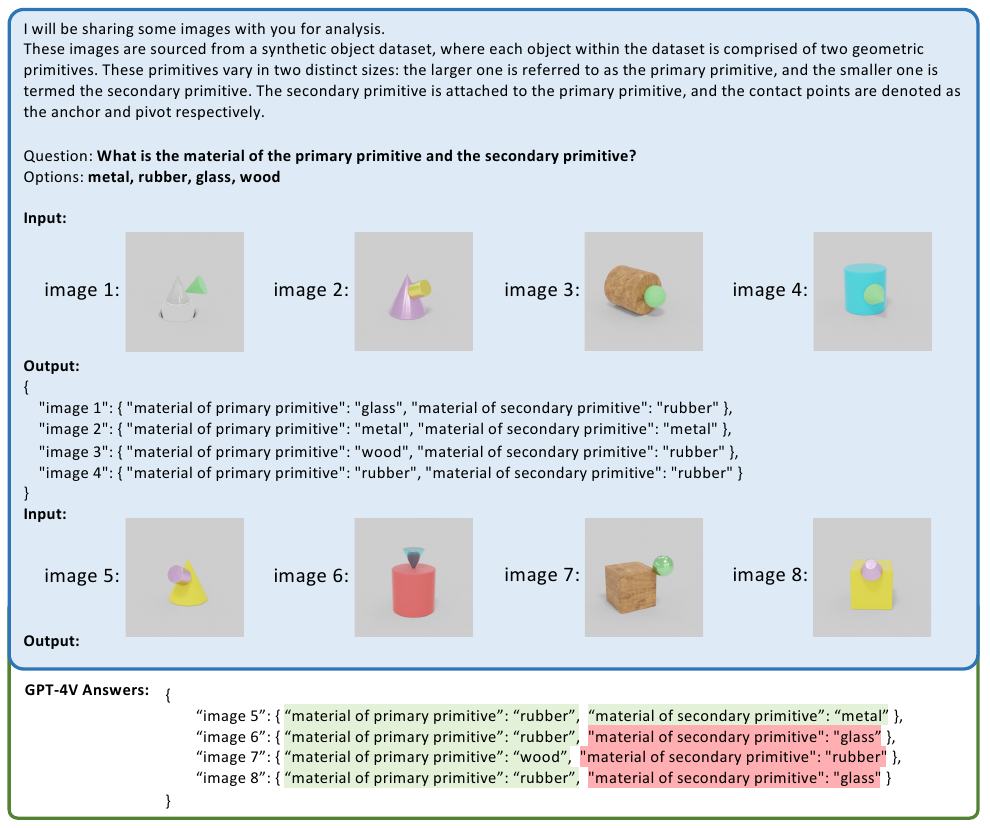}
\end{center}
\caption{The results of GPT-4V recognizing material when providing in-context examples.}
\label{fig:exp1-icl1}
\end{figure*}

\begin{figure*}[thb!p]
\begin{center}
\includegraphics[width=1.0\linewidth]{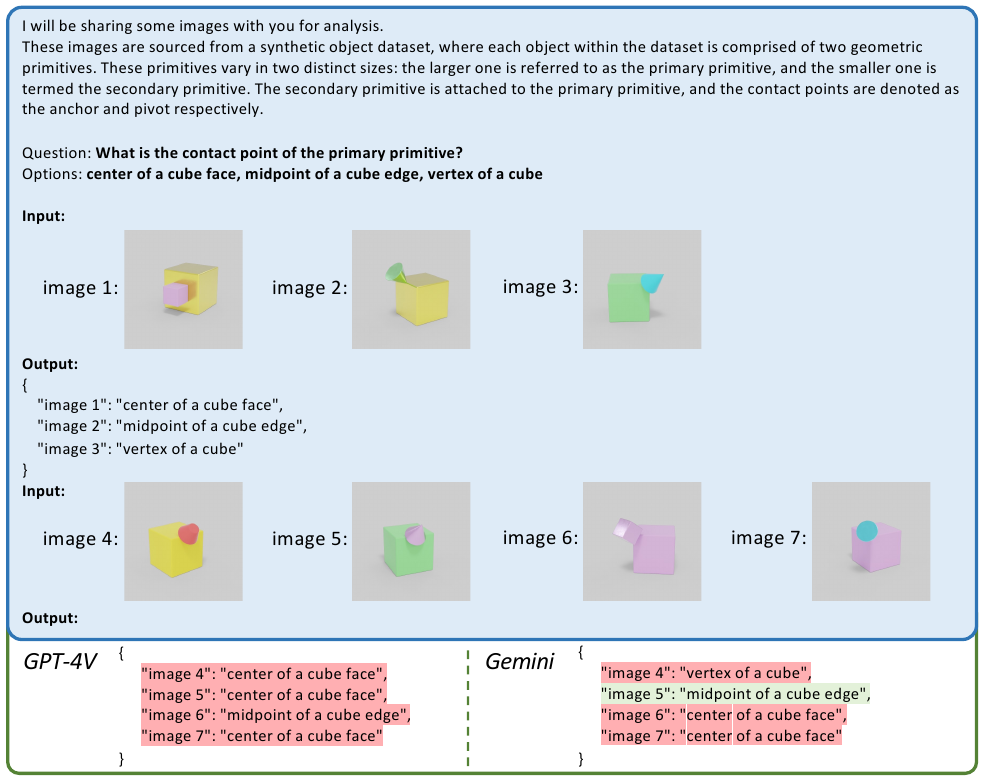}
\end{center}
\caption{The results of GPT-4V and Gemini recognizing contact point when providing in-context examples.}
\label{fig:exp1-icl2}
\end{figure*}

In Fig.\ref{fig:exp1-prompt-gpt}, Fig.\ref{fig:exp1-prompt-llava1}, and Fig.\ref{fig:exp1-prompt-llava2}, we illustrate some Pattern Perception results from four LMMs.
In Fig.\ref{fig:exp1-icl1} and Fig.\ref{fig:exp1-icl2}, we illustrate in-context learning results of GPT-4V and Gemini.

\subsection{Abstraction Alignment}

\begin{figure*}[thb!p]
\begin{center}
\includegraphics[width=1.0\linewidth]{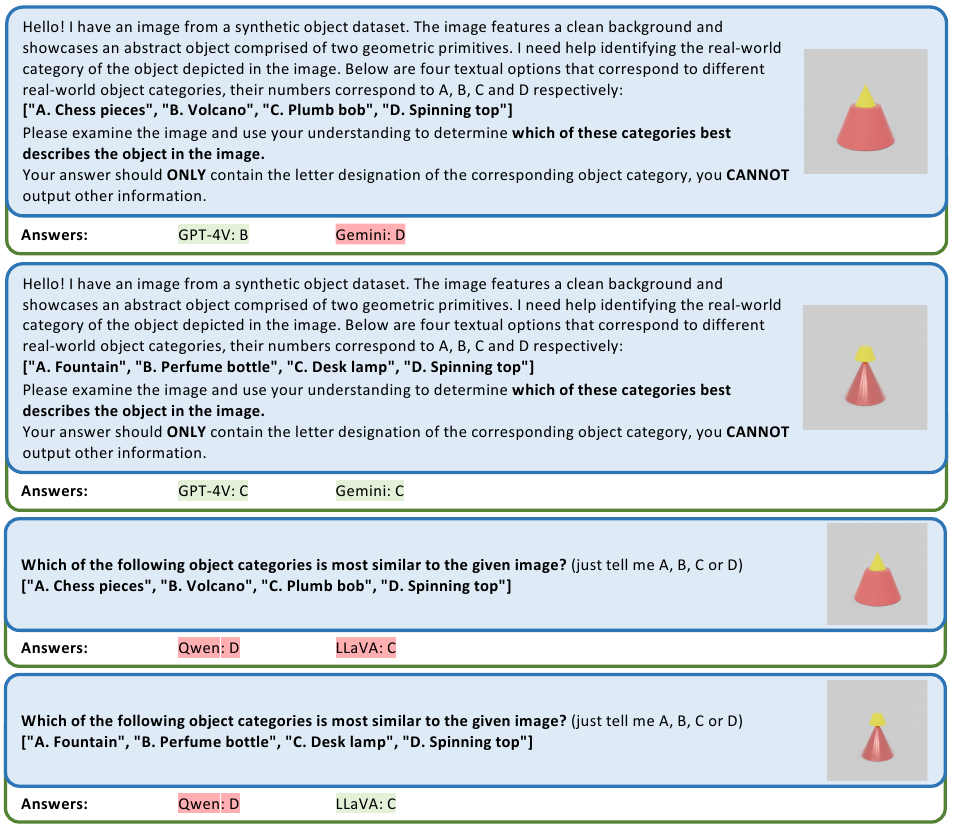}
\end{center}
\caption{Prompts for \textit{Img2Text} and outputs from all LMMs.}
\label{fig:exp2-1}
\end{figure*}

\begin{figure*}[thb!p]
\begin{center}
\includegraphics[width=1.0\linewidth]{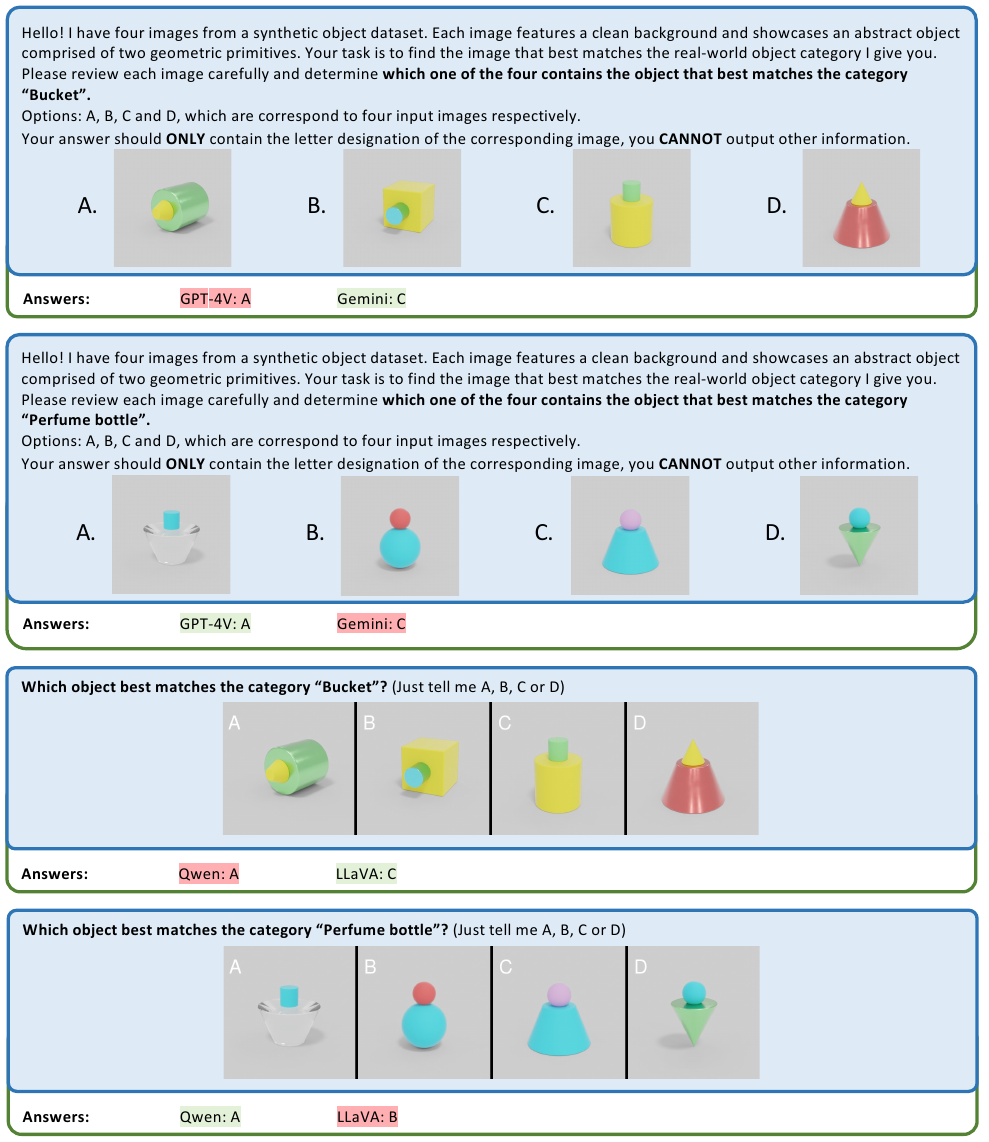}
\end{center}
\caption{Prompts for \textit{Text2Img} and outputs from all LMMs.}
\label{fig:exp2-2}
\end{figure*}

In Fig.\ref{fig:exp2-1} and Fig.\ref{fig:exp2-2}, we illustrate the prompts for \textit{Img2Text} and \textit{Text2Img} along with the outputs from all LMMs.

\subsection{Category Building}

\begin{figure*}[thb!p]
\begin{center}
\includegraphics[width=1.0\linewidth]{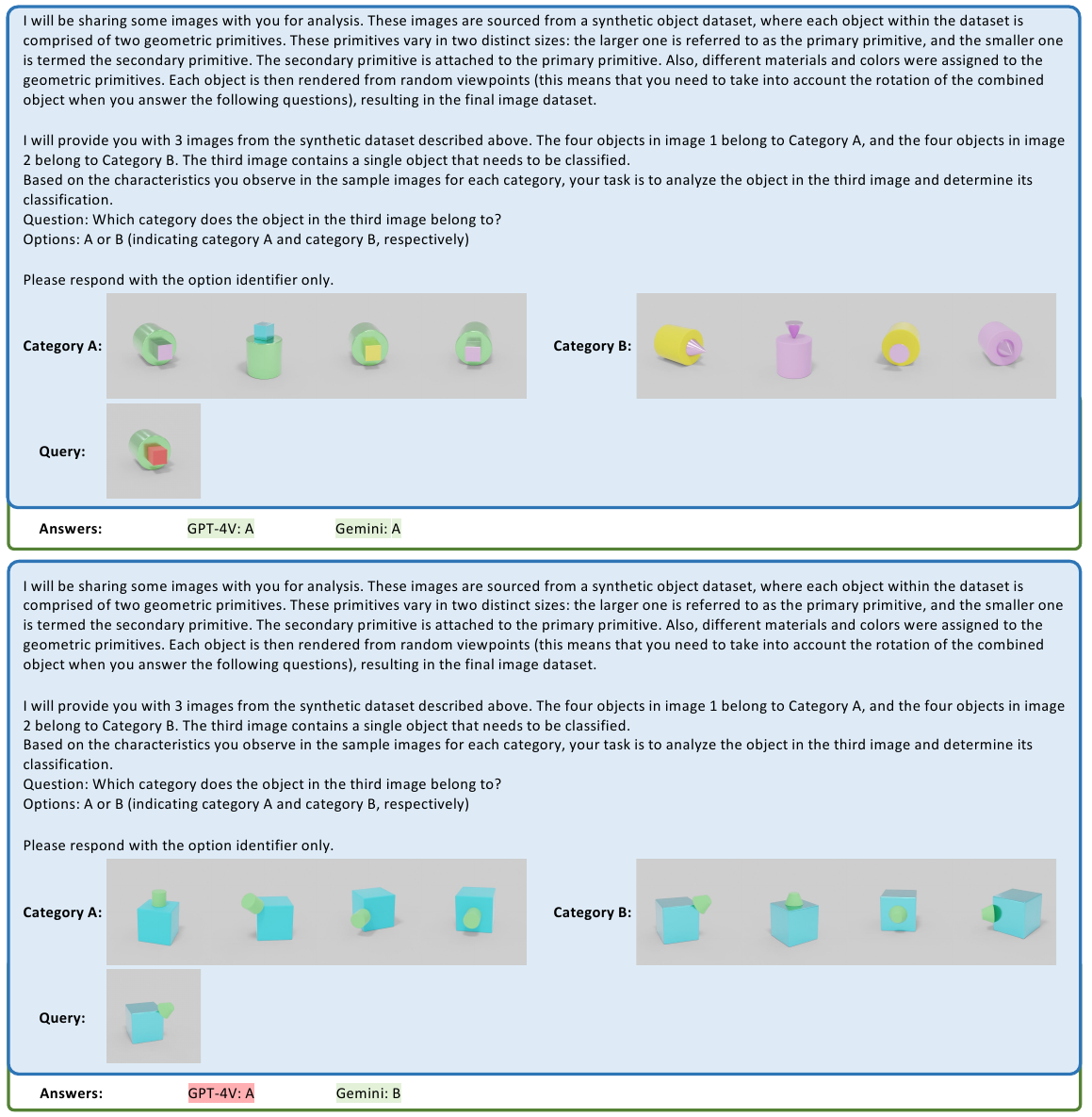}
\end{center}
\caption{Examples of the \textbf{Category Building} task, using standard prompts, showcasing outputs from GPT-4V and Gemini.}
\label{fig:exp3-gpt}
\end{figure*}

\begin{figure*}[thb!p]
\begin{center}
\includegraphics[width=1.0\linewidth]{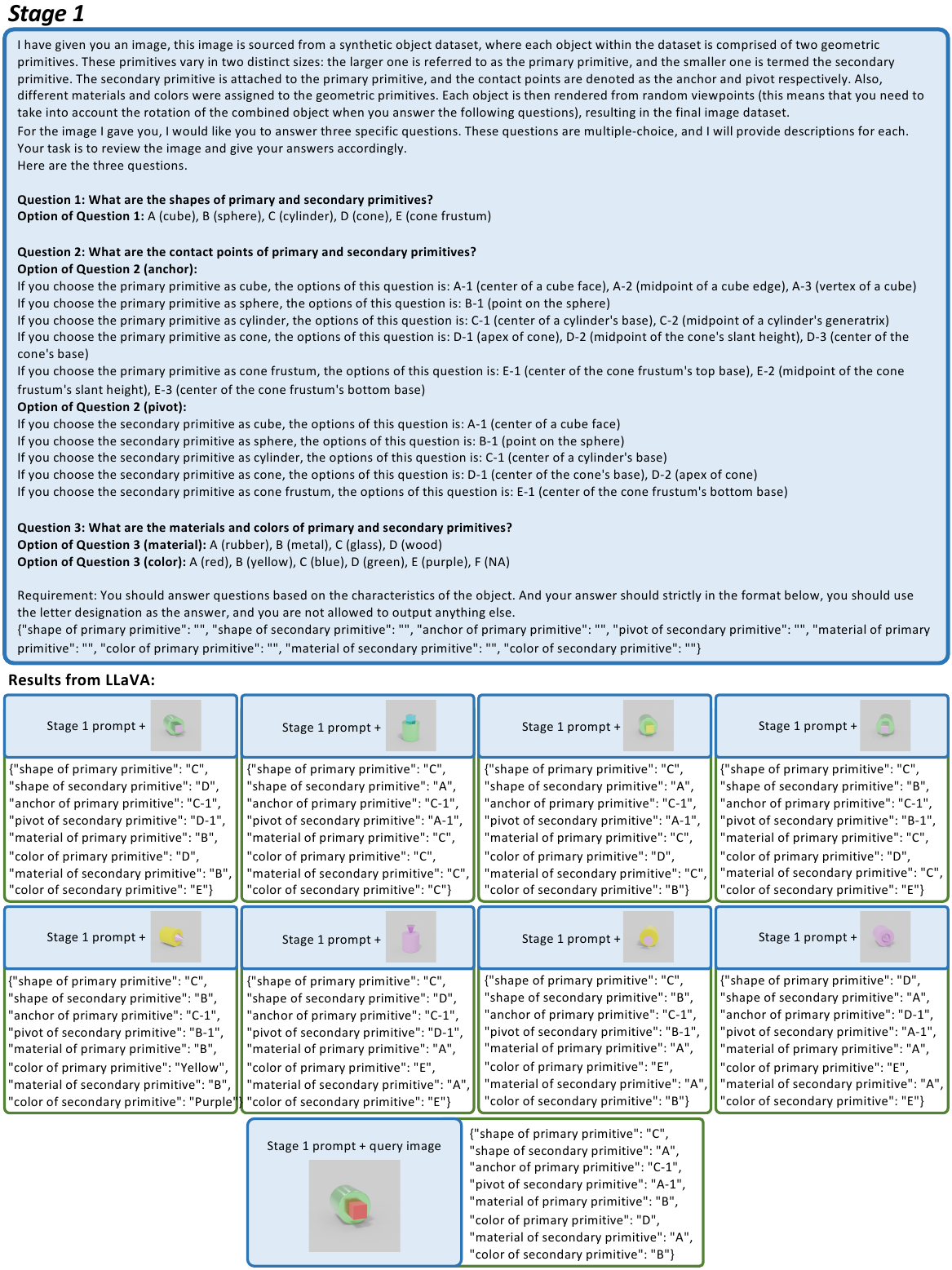}
\end{center}
\caption{An example of the \textbf{Category Building} task. In Stage 1, LLaVA answers eight questions regarding pattern perception for each image.}
\label{fig:exp3-llava1-1}
\end{figure*}

\begin{figure*}[thb!p]
\begin{center}
\includegraphics[width=1.0\linewidth]{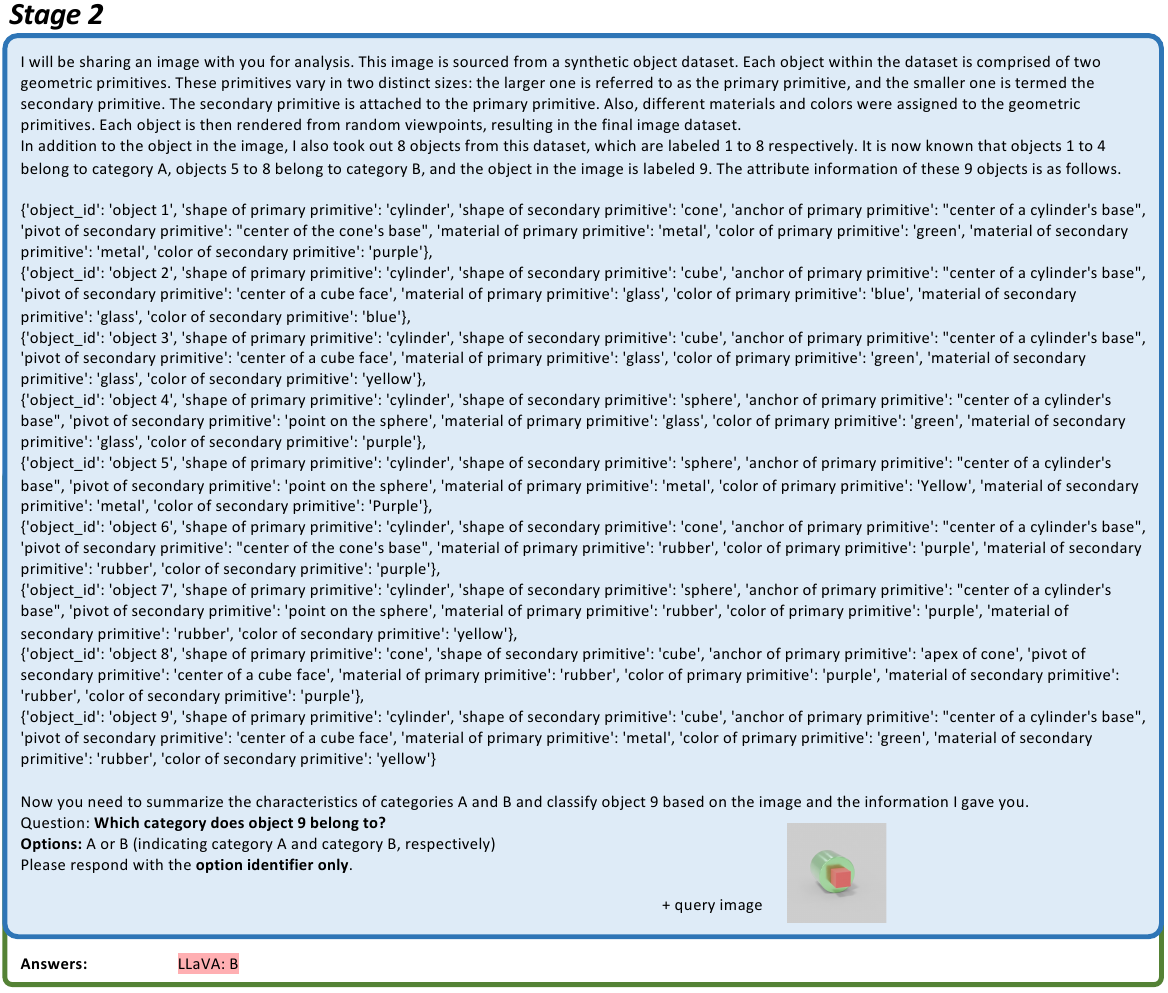}
\end{center}
\caption{An example of the \textbf{Category Building} task. In Stage 2, the analyses of all images are concatenated into the prompt, assisting LLaVA in classifying the query image.}
\label{fig:exp3-llava1-2}
\end{figure*}

\begin{figure*}[thb!p]
\begin{center}
\includegraphics[width=1.0\linewidth]{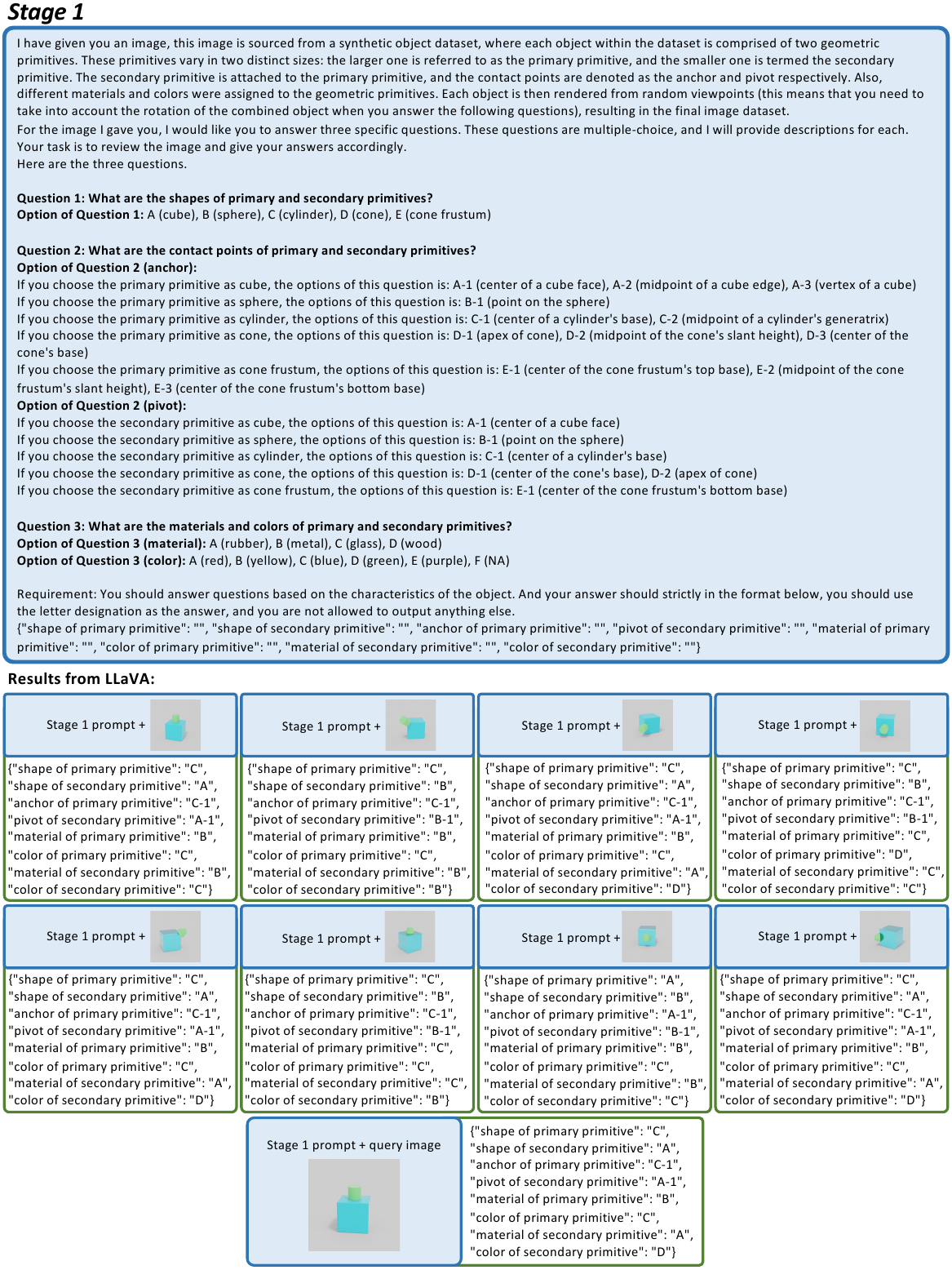}
\end{center}
\caption{An example of the \textbf{Category Building} task. In Stage 1, LLaVA answers eight questions regarding pattern perception for each image.}
\label{fig:exp3-llava2-1}
\end{figure*}

\begin{figure*}[thb!p]
\begin{center}
\includegraphics[width=1.0\linewidth]{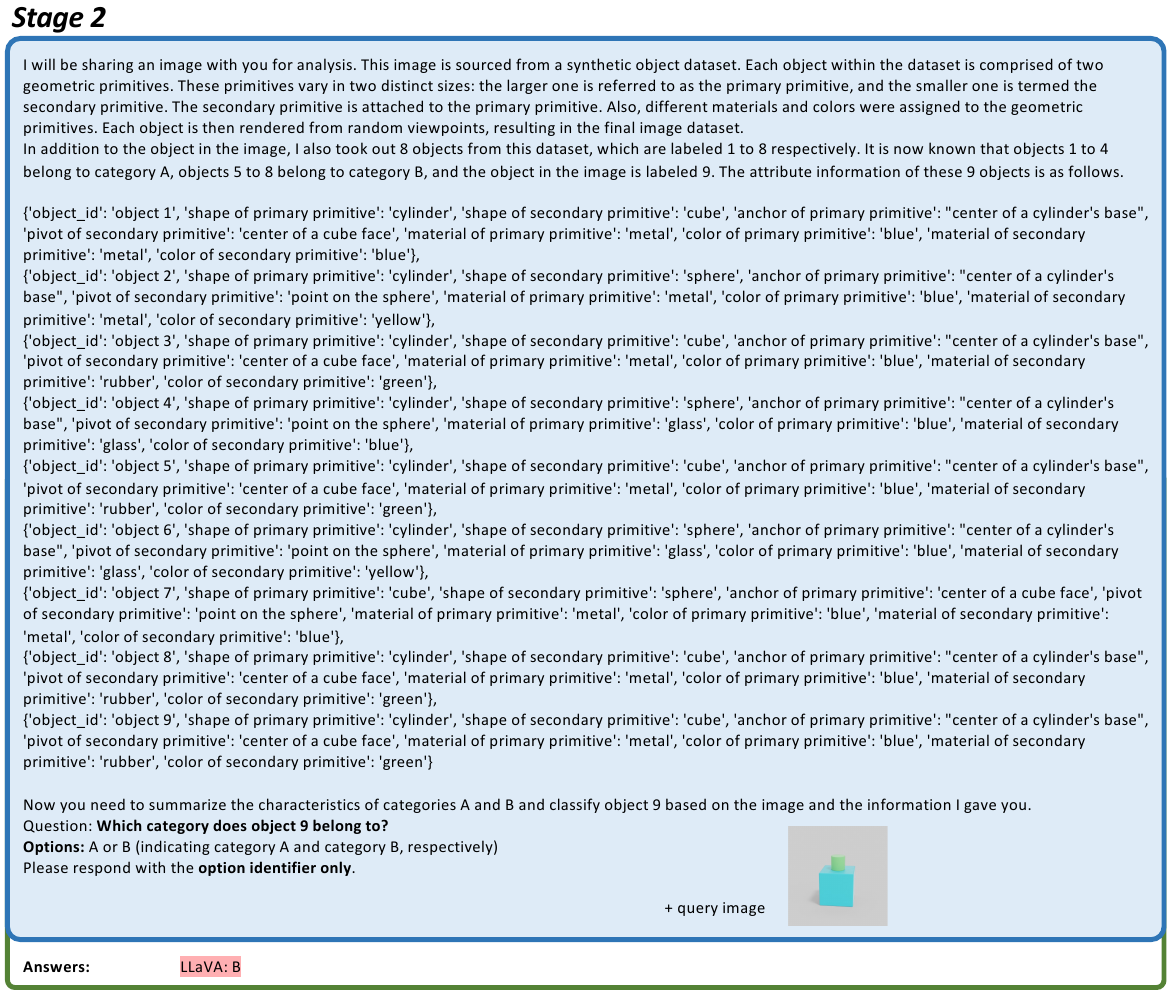}
\end{center}
\caption{An example of the \textbf{Category Building} task. In Stage 2, the analyses of all images are concatenated into the prompt, assisting LLaVA in classifying the query image.}
\label{fig:exp3-llava2-2}
\end{figure*}

We start by posing questions from the most natural prompts and gradually add hints or simplify the questions in the prompt when LMMs are completely unable to produce effective responses, until it is confirmed that LMMs can fully understand the question.
Samples of abstract categories are provided in groups of category A and category B.
Both GPT-4V and Gemini support multiple images as inputs.
Therefore, we combine four samples from category A into a large composite image, as well as four samples from category B, along with one image to be tested, forming a set of three images (4+4+1) per question. The prompt is presented in Fig.\ref{fig:exp3-gpt}.
LLaVA and Qwen do not support inference with multiple image inputs, therefore we design a manual chain of thought process, as shown in Fig.\ref{fig:exp3-llava1-1}-Fig.\ref{fig:exp3-llava2-2}.
Specifically, we require LLaVA to perform pattern perception on the nine input images, then concatenate the output results as input into the category building prompt, asking LLaVA to classify the test image according to the perceived patterns (in text) and the test image.
However, Qwen cannot even handle the perception questions for eight patterns simultaneously.
If we split the queries, it would require 72 queries to obtain the pattern perception results for all nine images, which is overly complex and cumbersome.
Therefore, we further simplify the process by not requiring Qwen to output the perception results in a specific format.
Instead, we allow it to perform free captioning on the input samples.
Subsequently, we concatenate the generated captions into the category building prompt and perform the final classification task based on the text input and the test sample.
Nevertheless, regardless of which instruction is chosen, Qwen is unable to produce valid results, or it answers all questions with the same option (for example, "A").
Therefore it is judged as unable to complete this task, and consequently, as described in \blue{footnote 2 of the main paper}, there is no corresponding bar in \blue{Fig.7 in the main paper}.

\begin{figure*}[thb!p]
\begin{center}
\includegraphics[width=1.0\linewidth]{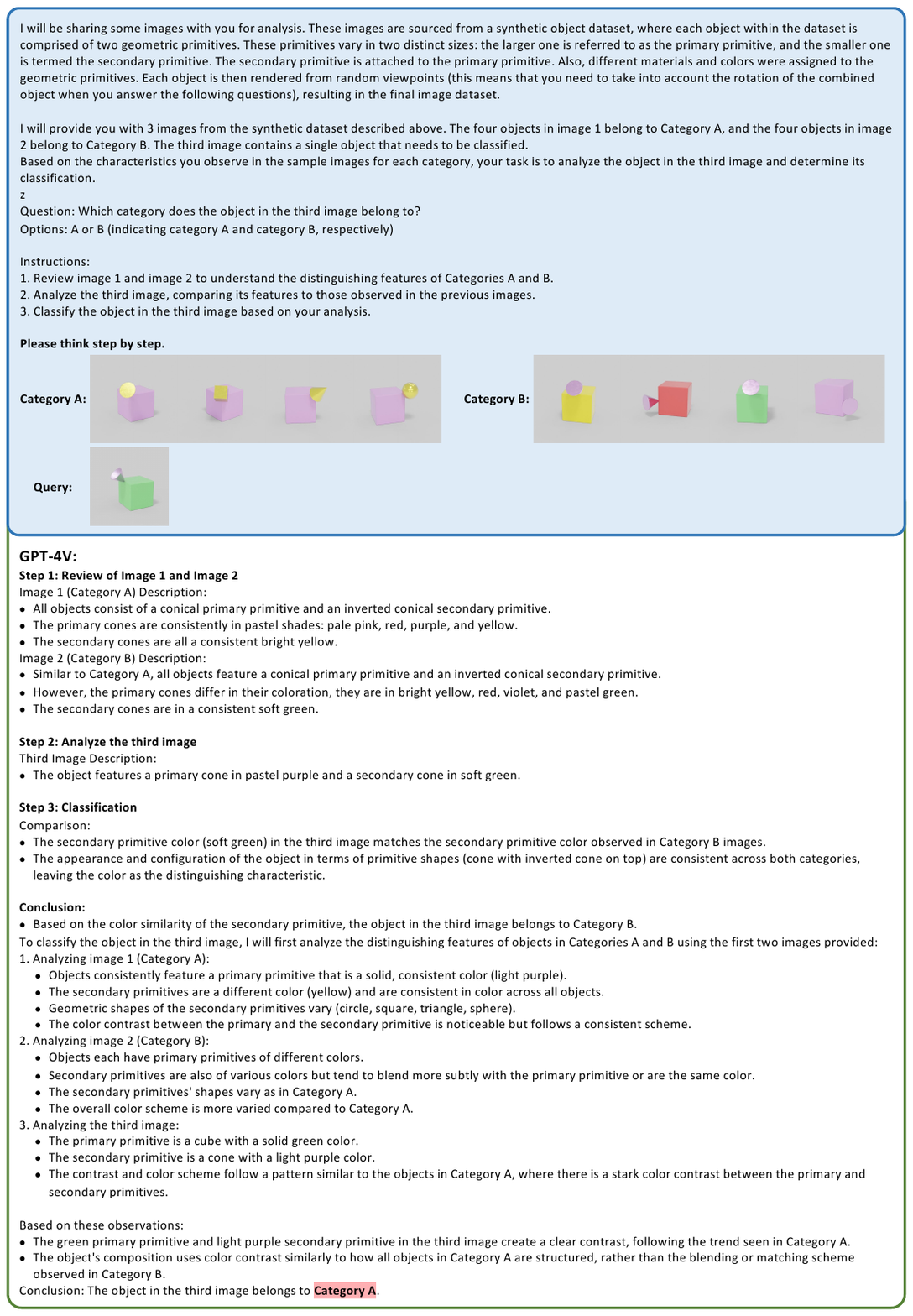}
\end{center}
\caption{An example from the task \textbf{Category Building} demonstrates GPT-4V's analytical process and classification result under the Chain of Thought (CoT) setting.}
\label{fig:exp3-gpt-cot}
\end{figure*}

\begin{figure*}[thb!p]
\begin{center}
\includegraphics[width=1.0\linewidth]{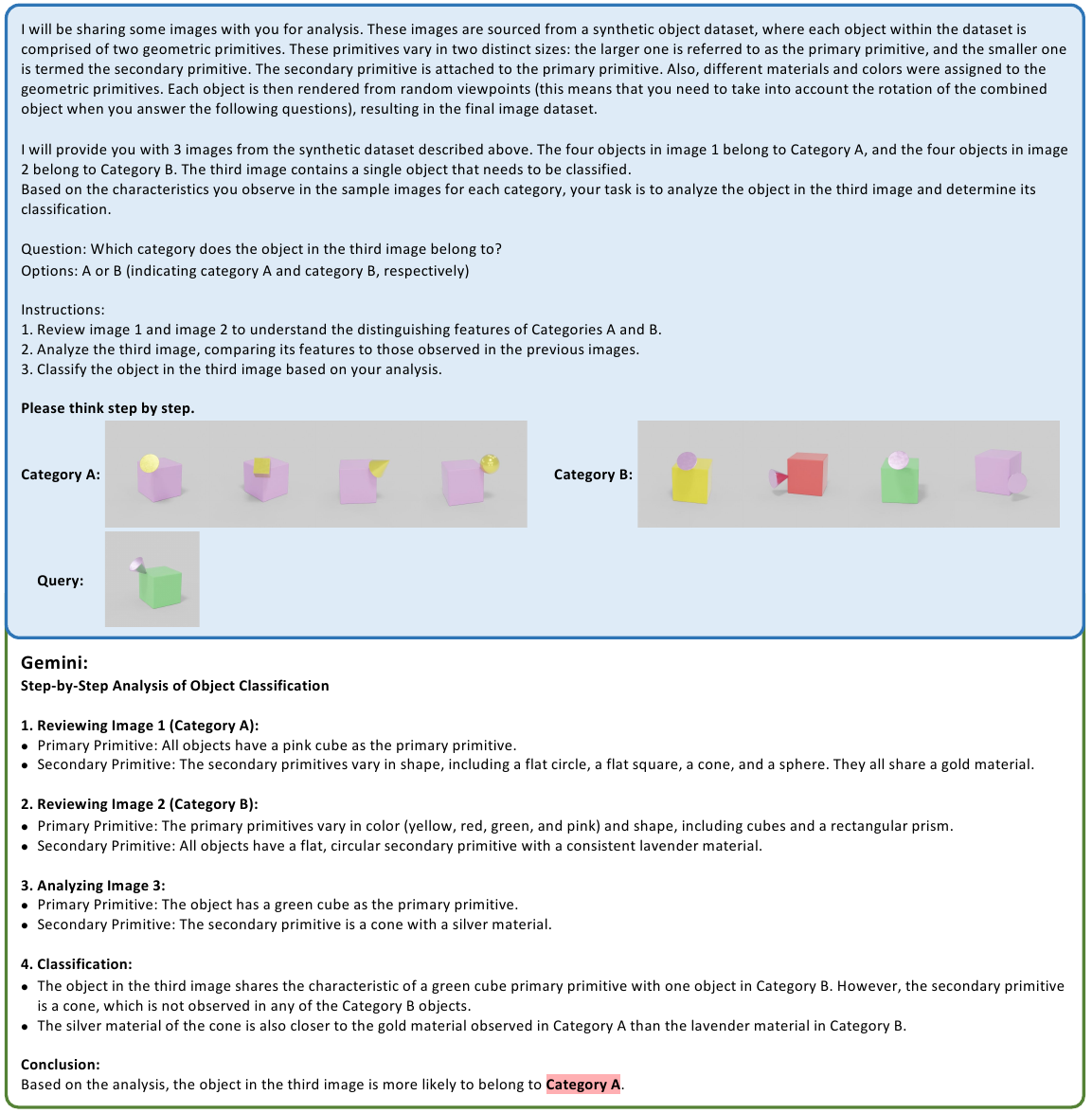}
\end{center}
\caption{An example from the task \textbf{Category Building} demonstrates Gemini's analytical process and classification result under the Chain of Thought (CoT) setting.}
\label{fig:exp3-gemini-cot}
\end{figure*}

We also illustrate chain-of-thought (CoT) results of GPT-4V and Gemini in Fig.\ref{fig:exp3-gpt-cot} and Fig.\ref{fig:exp3-gemini-cot}.

\section{More Discussion on Task Difficulty}
\label{sec:discussion}

As discussed in \blue{Sec.4.2 in the main paper}, we continue to focus on the \textbf{difficulty} of the ComBo benchmark in this section.
To verify that the questions in ComBo benchmark are much less difficult than these benchmarks in terms of logical reasoning, we use some smaller, commonly used computer vision models --- \textit{ResNet-50} and \textit{ViT-B/16} --- to complete the same tasks.
By compared with LMMs by pre-trained and fine-tuned models, readers can better perceive the \textbf{fundamental} characteristics of the ComBo benchmark.

In the most challenging task of Category Building, for LMMs that can accept natural language input, we describe the problem and directly ask questions in the form of \textit{instruction-following QA}, requiring them to respond in the specified format (usually in JSON format, as we do in the main paper).
For CNN and ViT, we simply extract features from the input images by the pre-trained models, and then calculate the average features of all samples in each abstract category as its prototype, and classify the test image according to the nearest principle.
This method is called \textit{clustering \& retrieval}.
As shown in the upper part of Table \ref{tab:category-building-finetuned}, through \textit{clustering \& retrieval}, ResNet-50 and ViT-B/16 both perform similarly to GPT-4V; in contrast, there is a significant gap between the pre-trained LLaVA and GPT-4V, which aligns with our prior.

\begin{table}[htbp]\small
    \begin{center}
    \begin{tabular}{cl|c|cccc}
    \toprule
    \multicolumn{2}{c|}{Model} & Method & Easy & Medium & Hard & Expert \\ \midrule
    \multirow{2}{*}{LMM} & LLaVA & \multirow{2}{*}{instruction-following QA}  & 0.48 & 0.38 & 0.36 & 0.26 \\
    & GPT-4V & & 0.98 & 0.92 & 0.80 & 0.72 \\ \midrule
    \multirow{2}{*}{pre-trained} & ResNet-50 & \multirow{2}{*}{clustering \& retrieval} & 0.98 & 0.80 & 0.70 & 0.58 \\
    & ViT-B/16 & & 1.00 & 0.98 & 0.86 & 0.66 \\ \midrule
    \multirow{2}{*}{fine-tuned} & ResNet-50 & \multirow{2}{*}{clustering \& retrieval} & 1.00 & 1.00 & 1.00 & 0.92 \\
    & ViT-B/16 & & 1.00 & 1.00 & 0.96 & 0.94 \\ \midrule
    fine-tuned & LLaVA & instruction-following QA* & 1.00 & 1.00 & 0.98 & 0.92 \\ \bottomrule
    \end{tabular}
    \end{center}
    \caption{Category Building results for LMMs, pretrained models, and fine-tuned models. Instruction-following QA marked with an asterisk (*) indicates that the fine-tuned LLaVA has lost most of its general visual capabilities, only able to complete Pattern Perception and Category Building well. Moreover, it is extremely sensitive to prompts and cannot tolerate minor perturbations.}
    \label{tab:category-building-finetuned}
\end{table}

To obtain a more appropriate feature extractor by fine-tuning CNN and ViT, we leverage the straightforward classification task among 9,504 categories as an auxiliary task.
After a quick fine-tuning for one epoch, as shown in the lower part of Table \ref{tab:category-building-finetuned}, applying the same \textit{clustering \& retrieval} methods can almost perfectly complete the task of Category Building, achieving an classification accuracy far higher than GPT-4V.
As a comparison, we also fine-tune LLaVA.
We construct a dataset containing 33,656 image descriptions and 40,000 task-specific question-answer pairs, and fine-tune LLaVA with LoRA. 
As shown in the last row of Table \ref{tab:category-building-finetuned}, fine-tuned LLaVA can also achieve excellent performance on this task, but at the cost of greatly compromising its general visual capabilities.

In this section, the fine-tuned CNN, ViT, and LLaVA can all be regarded as \textbf{specialized models for completing specific tasks}.
They perform excellently on the ComBo benchmark using pure visual capabilities, which is to be expected.
The above experiment demonstrates that the ComBo benchmark can effectively test the fundamental visual capabilities of LMMs in a Q\&A manner.
The experimental results indicate that current LMMs still have gaps compared to specialized models or humans when performing basic perception tasks.
We believe that a powerful LMM should pay more attention to the ability of fundamental visual perception while accomplishing various high-level tasks, so that LMMs can develop more robustly.

\newpage

\newpage

\bibliography{bmvc887}
\end{document}